%% file: acl_latex.tex
\pdfoutput=1
\documentclass[11pt]{article}

\usepackage[]{acl}
\usepackage{amsmath}
\usepackage{tablefootnote}
\usepackage{times}
\usepackage{latexsym}
\usepackage{booktabs,multirow}
\usepackage{bbding,pifont,adjustbox,amssymb}
\usepackage{colortbl}
\usepackage{xcolor}
\definecolor{Gray}{gray}{0.92}
\usepackage[T1]{fontenc}
\usepackage{listings}
\usepackage[utf8]{inputenc}

\usepackage{microtype}

\title{Generalized Quantifiers as a Source of Error \\in Multilingual NLU Benchmarks} 
\author{Ruixiang Cui, Daniel Hershcovich, Anders Søgaard\\
  University of Copenhagen \\
  \quad\texttt{\{rc, dh, soegaard\}@di.ku.dk}
 }

\begin{document}
\maketitle
\begin{abstract}
Logical approaches to representing language have developed and evaluated computational models of quantifier words since the 19th century, but today's NLU models still struggle to capture their semantics. We rely on Generalized Quantifier Theory for language-independent representations of the semantics of quantifier words, to quantify their contribution to the errors of NLU models. We find that quantifiers are pervasive in NLU benchmarks, and their occurrence at test time is associated with performance drops. Multilingual models also exhibit unsatisfying quantifier reasoning abilities, but not necessarily worse for non-English languages. To facilitate directly-targeted probing, we present an adversarial generalized quantifier NLI task (GQNLI) and show that pre-trained language models have a clear lack of robustness in generalized quantifier reasoning.
\end{abstract}

\section{Introduction}

Quantifier words---such as {\em each} or {\em most} or {\em more than three}---have been extensively studied, both in logic and in linguistics \citep{westerstaahl1989quantifiers,peters2006quantifiers}, going all the way back to \citet{frege1879begriffsschrift}. In this paper, we examine the extent to which they present a challenge to modern NLU systems. Our analysis is motivated by three  observations:

{\bf Quantifier words are abstract} Unlike nouns, verbs and adjectives, quantifier words do not have referents out in the world. Rather, quantifier words specify relationships between sets of entities, events and properties. To provide intuitions about the semantics of quantifier words, and to be able to refer to quantifiers in a language-independent way, we rely on the notion of generalized quantifiers \cite{mostowski1957generalization}, as described in \S2. 

{\bf Quantifier words vary across languages} Quantifier word inventories differ across languages. Often what is considered rough translation equivalents also differ in syntax, fine-grained semantics or pragmatics. \newcite{10.3389/fpsyg.2019.00957} show, e.g., that perceptions of the numerical bounds of existential quantifiers differ across speakers of English, French, Slovenian, and German. Other papers showing discrepancies between quantifier systems include comparisons of Salish to English \cite{matthewson2001quantification}, Adyghe to English \cite{Nikolaeva2012}, or of Dutch, Hebrew and Bengali \cite{10.2307/25001107}. The cross-linguistic differences in how generalized quantifiers are expressed motivates a cross-lingual error analysis, since quantifiers may contribute more to error when processing some languages rather than others. 

\input{tables/examples.tex}
\input{tables/categorization_set.tex}
{\bf Quantifier words are important} Quantifier words are extremely important for tasks that require inference, including natural language inference, question answering, fact-checking, etc. Datasets have, for example, been developed for numerical reasoning in English \cite{dua-etal-2019-drop}. Several researchers have  identified quantifier words as important sources of errors for natural language processing systems \cite{joshi-etal-2020-taxinli}; see Table~\ref{tab:examples} for examples of such errors. Unfortunately, most efforts have concentrated on subsets of quantifier words and on English. 

{\bf Contributions} We analyze how quantifiers are represented in NLU benchmarks, and how their occurrence at test time contributes to errors by neural language models (LMs). We derive a linguistically motivated 11-way categorization set for generalized quantifiers and look into their distribution in three steps: (a) monolingual NLI; (b) cross-lingual NLI; (c) cross-lingual question answering. We also propose GQNLI\footnote{\url{https://github.com/ruixiangcui/GQNLI}}, an adversarial generalized quantifier NLI challenge dataset. Our work shows that (i) generalized quantifiers are pervasive and cause overall performance drops in NLU benchmarks; (ii) the contribution of quantifier words to system error varies across languages; and (iii) generalized quantifiers are particularly difficult for LMs in interaction with negation and subsumption. 

\section{Background}

Generalized quantifiers (GQs) are developed upon first-order predicate logic, denoting relations between sets \citep{mostowski1957generalization}. Given a universe \textit{E}, a quantifier \textit{Q} would be treated as a mapping \(Q_{E}\) from the Cartesian product of powersets \(\mathcal{P}(E) \times \mathcal{P}(E)\) to the set \{\textit{false,true}\} or, as a binary relation on subsets of \textit{E} \citep{dvorak2015}. GQs are generalizations of the %universal 
\(\forall\),%and the existential 
\(\exists\) quantifiers from first-order predicate logic  \citep{mostowski1957generalization, perlindstrom1966first,montague1973proper,Bach1995QuantificationIN,Keenan2012HandbookOQ}. A generalized quantifier is, abstractly, a relation between sets. Generalized quantifier theory, while developed by logicians, is used by formal linguists to analyze the meaning of quantifier words in combination with referential expressions 
\citep{barwise1981generalized,higginbotham1981questions}.

Most human languages contain ways of expressing generalized quantifiers, and their semantics exhibit striking similarities across languages \cite{matthewson2004methodology, 2008universals,SteinertThrelkeld2019LearnabilityAS}. At the same time, generalized quantifiers can be instantiated very differently across languages due to pragmatic considerations \citep{grice1989studies} or cognitive economy
and cost-benefit optimisation in the exchange of information \citep{levinson2000presumptive,e23101335,10.1162/ling_a_00461}. Quantifier words also exhibit syntactic differences, e.g., with some languages having specialized words
to express quantity, while others rely on metaphorical usage of common nouns \citep{katsos2012acquisition}.  In English, {\em most} is a determiner, but Spanish and French express the same concept through common nouns, {\em la mayor\'{i}a} and {\em la majorit\'{e}}. The relative stability of the core semantics of quantifiers makes a cross-linguistic comparison possible, but the syntactic and pragmatic variation associated with the expression of generalized quantifiers poses a challenge for multilingual NLU.    
We consult quantifier taxonomy studies \citep{Keenan1997GeneralizedQI, peters2006quantifiers, szymanik-thorne-2015-semantic, Szymanik2016} and derive a categorization set for quantifier analysis in NLU benchmarks. In Table~\ref{tab:categorization_set}, we list the 11-way quantifier categorization set and their logical denotation based on set theory. 

While other foci of formal linguistics have attracted the attention of NLP researchers---including coreference \citep{ws-2019-models, crac-2020-models}, negation \citep{hossain-etal-2020-analysis,hartmann-etal-2021-multilingual}, 
and consistency \citep{li-etal-2019-logic,ribeiro-etal-2019-red,asai-hajishirzi-2020-logic,10.1162/tacl_a_00450}---there has been little work on generalized quantifiers as a source of error in NLU, let alone in multilingual NLU. It remains an open problem whether LMs represent the semantics of quantifiers words adequately, or if they provide a basis for resolving scopal ambiguities.\footnote{Note that generalized quantifiers are not always {\em explicit} in discourse. The sentence {\em inadequate sleep causes obesity} should be interpreted as {\em Most of those who do not sleep adequately, gain weight} \citep{zadeh1983computational}. Such implicit quantifiers related to pragmatic variation are important for language understanding, but will be ignored in this work.}

\section{NLU Benchmarks}

We conduct an error analysis focusing on the role of  generalized quantifiers in %NLU benchmarks in the following steps. We regard 
two %types of 
NLU tasks, Natural Language Inference (NLI) and Question Answering (QA), which generally %as of interest,  for they particularly 
require understanding of quantifiers. For each type of task, both monolingual and cross-lingual evaluation are conducted. We focus on generalized quantifiers in the {\em hypotheses} in NLI examples---and on generalized quantifiers in the {\em question} fields in question answering. 
\input{tables/task_stats_nli.tex}
To this end, we identify quantifiers by the lemma and the universal dependency relation \citep{nivre-etal-2020-universal} of a quantifier after preprocessing the sentences using \textit{Stanza} \citep{qi-etal-2020-stanza}. Take the sentence ``The Yiddish culture has survived for more than a thousand years.'', we annotate it as ``The/\textit{det} Yiddish/\textit{amod} culture/\textit{nsubj} have/\textit{aux} survive/\textit{root} for/\textit{case} more/\textit{advmod} than/\textit{fixed} a/\textit{det} thousand/\textit{nummod} year/\textit{obl} ./\textit{punct}''. By matching the regex pattern of the quantifier ``more than k'', in this case \textit{``((more|great)\symbol{92}/advmod than\symbol{92}/(fixed|case)|at\symbol{92}/case least\symbol{92}/nmod) .+\symbol{92}/nummod .+\symbol{92}/(nsubj|obj|obl)''}, we approximate the surface form of the type ``more than k''.Through matching quantifier patterns, we are able to find entries in which quantifiers are instantiated. See Appendix~\ref{app: Regular Expressions} for the list of regex patterns we write to identify GQs. In Table~\ref{tab:task_stats_nli} and Table~\ref{tab:task_stats_qa}, we present the statistics of the quantifier distributions in NLI and QA tasks, respectively. As can be seen, quantifiers are indeed widespread in NLU tasks, accounting for roughly 10\% in NLI tasks and 5\% in QA tasks.
% Although the distribution trend is not identical; 
We will further discuss the statistics and experiments in the following section.

\section{Quantifiers in English NLI Benchmarks}
\label{sec:Quantifiers in Four English NLI Tasks}

\begin{figure}[!t]
    \centering
    \includegraphics[width=\columnwidth]{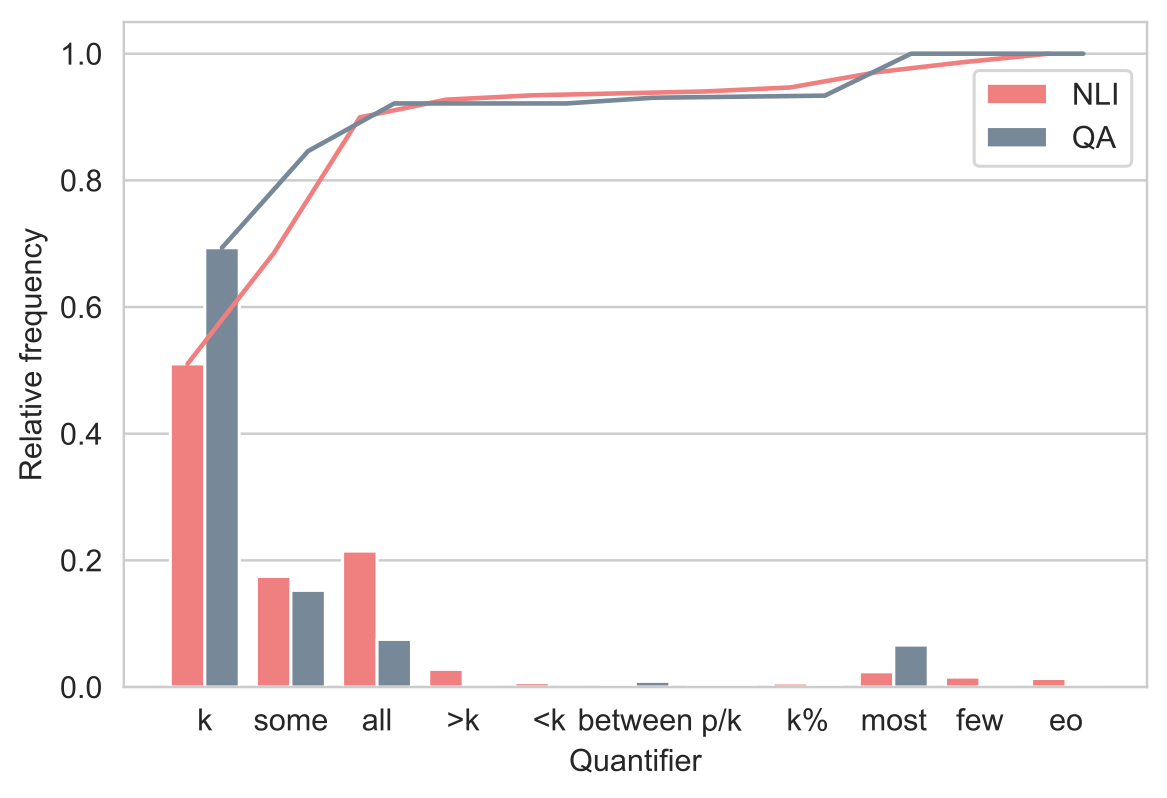}
    \caption{Relative distribution of quantifiers in NLI and QA tasks ranked by semantic complexity. The bars show the relative frequency of such quantifier and the lines indicate the cumulative frequency for a task. }
    \label{fig:frequency}
\end{figure}
\input{tables/nli_result.tex}

NLI is commonly framed as a three-way classification task with labels \textit{entailment}, \textit{contradiction} and \textit{neutral} \cite{bowman-etal-2015-large}. While SOTA models exhibit low error rates on NLI benchmarks, it is unclear when they succeed or fail in their underlying reasoning. We are interested in whether generalized quantifers challenge modern NLI models. In our error analysis, we initially focus on three English NLI datasets, MultiNLI \citep[MNLI;][]{williams-etal-2018-broad}, SNLI \citep{bowman-etal-2015-large} and ANLI \citep{nie-etal-2020-adversarial} as testbeds. 

Table~\ref{tab:task_stats_nli} presents statistics of quantifier distribution in these datasets, where we observe that, across, about 10\% of all hypotheses contain quantifier words, indicating the pervasiveness of quantification. We also plot the frequency of quantifiers in NLI in Figure~\ref{fig:frequency} and find the quantifier word distribution follows Zipf's law \citep{Zipf1949HumanBA}. Note the top three most common quantifiers account for more than 90\% of all. 

\paragraph{Experiments and Results}
In order to investigate whether NLU systems can solve quantifiers in NLI, we experiment with two pretrained LMs: BERT\footnote{\texttt{wwm\_cased\_L-24\_H-1024\_A-16}} \citep{devlin-etal-2019-bert} and RoBERTa\footnote{\texttt{roberta-large}} \citep{Liu2019RoBERTaAR}. We use the codebase by \citet{nie-etal-2020-adversarial}. The training data combines SNLI, MNLI, FEVER-NLI \citep{nie2019combining} and ANLI. 

In Table~\ref{tab:nli_result}, we report the test set performance on SNLI and ANLI, and the dev set performance on MLNI \textit{matched} and \textit{mismatched} sections. We can observe that SOTA models suffer from performance drops across almost all quantification phenomena in every task. When it comes to performance over all quantifiers, the improvement from RoBERTa to BERT (2.2\%) is less prominent than that over full datasets (2.9\%), suggesting RoBERTa is particularly challenged. 

Taking a closer look at error by category, proportional quantifiers seem harder to solve than Aristotelian/counting quantifiers. Except for {\em k\%}, all proportional quantifiers---{\em p/k}, {\em most}, and {\em few}---are about 10\% lower than the five counting quantifiers (except {\em less than k}) with BERT; and about 5\% lower with RoBERTa. RoBERTa is not generally superior to BERT; e.g., for {\em k\%}, BERT outperforms it by 22\%. We show a pairwise analysis of how GQs affect performance when they appear in both the premises and hypotheses in the Appendix \ref{app:pairwise}. Generally, our results attest to the difficulty of resolving GQs in NLI benchmarks.

\section{Quantifiers in Cross-lingual NLU Benchmarks}
Quantifiers are acquired in similar orders across languages \cite{Katsos2016CrosslinguisticPI}, although languages express quantifiers in different ways. For example, there are eight different universal quantifiers with different level of distributivity in Malagasy \citep{Matthewson2008QuantificationAC}. This poses challenges to training multilingual LMs and transfer learning. We are interested in whether quantifiers are universally and evenly challenging for all languages. 

\input{tables/xnli_result.tex}

\paragraph{Quantifiers in Cross-lingual NLI}
We choose XNLI \citep{conneau-etal-2018-xnli}, a manual translation of the development and test set of MNLI into 15 languages, for this multilingual error analysis. We should clarify that for XNLI, the authors annotate entailment labels for the English data only and apply them to the other languages. We do not assume label changes due to translation in this study, but it is worth investigate in the future. We choose five languages belonging to different language families, namely Arabic, Chinese, German, Spanish and Vietnamese as targets. 
The last column in Table~\ref{tab:task_stats_nli} shows the numbers of quantifiers in XNLI. The distribution rate is 10\%. Note that the universal quantifier is the most common quantifier in XNLI. 

We fine-tune mBERT\footnote{\texttt{multi\_cased\_L-12\_H-768\_A-12}} \citep{devlin-etal-2019-bert} and XLM\footnote{\texttt{xlm-mlm-100-1280}} \citep{Lample2019CrosslingualLM} on the MNLI training set and evaluate them on XNLI. We report the results in Table~\ref{tab:xnli_result}. We find that performance varies across languages. For Chinese and Vietnamese, we see significant drops in performance for examples with GQs, whereas for Arabic and German, we see improvements. The results {\em per} quantifier are more homogeneous, however. 

Similar to our results for English, we can see that the lowest accuracies in XNLI are with proportional quantifiers, such as {\em most} and {\em few}. But the gap in non-English languages is wider for these two categories, especially for Chinese, the difference reaches 30\%. Other hard quantifiers include {\em all}, $>k$, $<k$, and {\em each other}.

\paragraph{Quantifiers in Cross-lingual QA}
\input{tables/task_stats_qa.tex}
\input{tables/xqa_result_mbert.tex}
Cross-lingual question answering (XQA) is another important  NLU task that evaluates the cross-lingual transferability of LMs. We evaluate the effect of quantifiers on system errors across two XQA datasets, namely XQuAD \citep{artetxe-etal-2020-cross} and MLQA \citep{lewis-etal-2020-mlqa}.
As demonstrated in Figure~\ref{fig:frequency}, quantifier word distributions in XQA tasks also follow Zipf's law, as in NLI tasks, but {\em k} is more frequent (perhaps because of a traditional emphasis on numerical reasoning), and we see less variance across languages. This is probably because question answering is targeting quantification less directly. 
To evaluate cross-lingual QA performance on GQs, we fine-tune mBERT and XLM-R\footnote{\texttt{xlm-roberta-large}} \citep{conneau-etal-2020-unsupervised} using \citet{hu2020xtreme}'s architecture. We present results for mBERT in Table~\ref{tab:xqa_result_mbert}; for XLM-R results, please refer to Appendix~\ref{app:xqa_result}. 

Just as with XNLI, LMs suffer from performance drops across all languages for almost all GQ phenomena with significant, cross-lingual variation. The most distinguished is that Exact Match (EM) suffers from a greater deterioration than F1 scores for all languages. For example, the weighted EM difference for mBERT on MLQA is 2.9\% while the weighted F1 is 1\%. As one example in Table~\ref{tab:examples}, we observe that the plausible answers selected by models, while being incorrect, result in a sharper decrease of EMs comparing to F1s. Questions containing GQs also tend to have less verbal answers comparing to those without GQs, and therefore require higher precision. 

Regarding cross-lingual comparisons, Chinese and Arabic are the two languages that do not have lower performance over GQs compared to the performance over the complete dataset. Despite the overall trends, subtle differences from XNLI performance still exist. For example, XLM-R is worse than mBERT on quantifier reasoning on XQuAD Chinese, especially at proportional quantifiers, but this is not the case on MLQA Chinese.

\section{GQNLI}
\input{tables/gqnli}
We have seen how quantifiers present challenges to NLI and QA models. 
Using an approach similar to ANLI \citep{nie-etal-2020-adversarial} and DynaBench \citep{kiela-etal-2021-dynabench}, we use model difficulty (RoBERTa's) as a heuristic to select hard examples for a challenge dataset that can hopefully be used to evaluate any future progress on this. 
%Although current NLU benchmarks test LMs' generalized quantifier reasoning to some extent, the variety and the difficulty are limited. In particular, quantifier scope ambiguity and the interaction with negation and coreference are rarely focused on. 
We propose GQNLI, a generalized quantifier NLI challenge dataset, consisting of 30 premises and 300 hypotheses. The average sentence lengths of hypothesis and premises are 15.97 and 7.35, respectively. Both numbers are comparable to those of MNLI, but lower than ANLI's \citep{williams2020anlizing}. It should be noted that  GQNLI is designed for evaluating future models; obviously not for benchmarking RoBERTa.

\paragraph{Dataset Creation}
Firstly, we manually create 100 premise-hypothesis pairs, in which various types of GQs appear. For each premise and hypothesis, the number of GQs varies from one to three. 
To choose the premises, we randomly sampled 100 premises with GQs from SNLI and ANLI test sets, respectively, and selected 10 premises in total, that we consider are semantically adequate for adding GQs and making simple hypotheses.

To construct the hypotheses, we rely on 
RoBERTa fine-tuned on MNLI %, to 
and manually select examples about which the model is unsure or incorrect. To focus on GQs, we keep the challenge examples otherwise simple \cite{ribeiro-etal-2020-beyond}, and avoid lexical variations in the hypotheses. Hard examples were found to be characterized by (i) mixing generalized quantifiers with other logical operators, such as subsumption or negation, and (ii) combining multiple different generalized quantifiers. 
We discuss these observations in Section \ref{sec:discussion}.

Two of the authors annotated the examples. The inter-annotator agreement (Fleiss' kappa) was 0.895, substantially higher than ANLI's (0.672--0.740). It is worth noting that the level of  semantic or pragmatic interpretation difference of GQs is reflected in the measurement. 

We augmented the examples by substituting non-quantifier words (e.g., replacing ``dogs'' with ``cats'') while keeping the labels, to exclude the effect of specific lexical items. The resulting labels are uniformly distributed. Table~\ref{tab:gqnli} presents GQNLI statistics. Since the dataset is curated to probe the ability to reason with quantifiers, the distribution of generalized quantifiers does not follow Zipf's law; see \S\ref{sec:Quantifiers in Four English NLI Tasks}. A list of GQNLI examples per category is shown in Appendix \ref{app:gqnli_example}.

\paragraph{Experiments and Results}
We evaluate seven types of models on GQNLI, fine-tuned with different combinations of NLI datasets. As data creation only relied on RoBERTa and MNLI, nothing prevents that models with different architectures and training data will perform well. They do not, however. The results are shown in Table~\ref{tab:gqnli}. 

We see that all models have great difficulty with GQNLI. With more training data, models improve, but the best performance is 48\%, less than 15 points above chance level. In general, the counting quantifiers, especially the existential and universal quantifiers, are easier than proportional quantifiers. Particularly, most models struggle with {\em less than k} and {\em between}. This is in some contrast with the NLU tasks studied above, where these quantifiers were among the easiest.

We also observe unstable GQ reasoning ability in simple word substitution cases. For instance, it happens for DeBERTa fine-tuned with M, F, Ling, DocNLI that it predicted correctly the contradiction relation between ``There are six children standing on top of a yellow mountain. Two thirds wear red tops and one third wear green.'' and ``Between 80\% and 90\% children do not wear red tops.'', but incorrectly when ``red'' is substituted  with ``beige'' and ``green'' with ``cyan''. We are yet to study what kind of cues lead to the instability. Our experiments suggest a lack of testing proportionality reasoning and robustness in existing benchmarks.

\section{Discussion}
\label{sec:discussion}
\paragraph{Negation}
The interaction between negation words and quantifiers increases semantic complexity \citep{10.2307/25000447,Horn2010TheEO}. We investigate whether this holds for NLI tasks, using negation cue detection to find all cases where a negation word and a quantifier appear in the hypotheses. 

We break down the performances on negation of the seven models in Appendix \ref{app:negation}. As indicated, LMs overall have polarized results for negation cases comparing to the entire dataset. We can see a majority of the models even predicted opposite labels for some GQ categories, with 0\% accuracy. BART is no longer the second best model, replaced by RoBERTa. The improvement by training with more data is overall consistent for reasoning over GQs with negation. 

For a cross-lingual investigation of the interaction of GQs and negation, we find that in XNLI, the number of cases combining both phenomena is insufficient: we identified four such cases, involving only the quantifiers ``all'' and ``more than.'' For English, mBERT predicted two cases successfully. For Chinese, German, Vietnamese and Arabic, one is correct. For Spanish, all are wrongly predicted.

It is evident that NLU models suffer from reasoning difficulties in certain cases when negation interacts with GQs, especially in cross-lingual evaluation. In future work, we are interested in expanding GQNLI to more instances and more languages to facilitate qualitative investigations.

\paragraph{Subsumption}
In generalized term subsumption languages \citep[TSLs;][]{Yen1991GeneralizingTS,ali1993natural}, a term \(a\) subsumes another term \(b\) if and only if the extension of \(a\) is a superset of the extension of \(b\) . Rather than surface number comparison, subsumption reasoning requires knowledge of the relations between supersets and subsets. For example, to decide whether ``There are six dogs. Three brown dogs, a black dog and a white dog run along the green grass'' entails ``One dog sits'', LMs should be aware that ``six dogs'' is a superset of the extension of the ``brown dogs'', ``black dog'' and ``white dog''. Another example in GQNLI is to infer whether ``There are twelve singers on a stage, less than half from Argentina and one from Cape Verde'' entails ``Several singers do not come from Chile''. 

We annotate 63 cases out of the first 100 in GQNLI requiring subsumption reasoning. 
We show the statistics and results regarding subsumption in Appendix \ref{app:Subsumption}. It can be seen that more training data leads to higher accuracies. Especially, DeBERTa fine-tuned with DocNLI, which unifies the two classes “neutral” and “contradict” into a new class “not entail”, has a significant improvement on subsumption cases with neutral label. The training bias give an advantage to the model on the subsumption subset, half cases of which are labelled neutral. But such bias has a negative effect on non-subsumption cases; the accuracy drops by 20.2\% comparing to the model without training with DocNLI. It is worth investigating whether DocNLI is truly helping subsumption reasoning in future work. Subsumption is a key concept in the study of knowledge representation \citep{Woods1991UnderstandingSA}, but is neglected in current NLP research. The fact that LMs struggle to perform subsumption reasoning asserts the necessity to explicit tackle the problem.

\section{Related Work}

We examine the sensitivity of NLU models to generalized quantifiers. These models are designed to induce correlations from large volumes of data, not to reason symbolically with logical quantifiers. Such models have, nevertheless, been probed for logical knowledge.

\citet{mul2019siamese}, for example, show neural networks encode fragments of first-order logic and exhibit zero-shot generalization ability. \citet{evans2018can} present a neural architecture that improves performance on propositional logical inference. \citet{bowman-etal-2015-recursive} also suggest neural networks learn semantic representations for logical inference in natural languages. However, on the same task, \citet{veldhoen2017can} find neural networks fail to do so on a more stringent test. \citet{geiger-etal-2019-posing} also show that neural networks fail to exhibit robust logical inference. \citet{srivastava-etal-2018-zero} use semantic parsers to encode quantifiers and improve zero-shot learning in classification tasks. \citet{haruta-etal-2020-logical} present a system that computes logical inference over GQs and see improvements on two specialized datasets, FraCaS \citep{cooper1994fracas} and MED \citep{yanaka-etal-2019-neural}. None of these papers explicitly discussed generalized quantifiers, and all were limited to studying the ability of neural networks to capture the logical semantics of English. 

Many studies have instead focused on LMs' ability to capture negation \citep{gururangan-etal-2018-annotation,naik-etal-2018-stress,hossain-etal-2020-analysis,ettinger-2020-bert,hartmann-etal-2021-multilingual} or coreference \citep{ye-etal-2020-coreferential,varkel-globerson-2020-pre,abdou-etal-2020-sensitivity}. %ruixiang{citation. are there more perspectives?}. \citet{ribeiro-etal-2020-beyond} introduced \texttt{CheckList} to test linguistic capabilities but none of them give relevance to quantifiers. A close field to qunatifier words is testing language models' numeric reasoning skill. 
Others have focused on LMs' ability to reason with numbers \citep{johnson-etal-2020-probing}. DROP \citep{dua-etal-2019-drop}, for example, is a question answering dataset designed specifically to probe LMs' ability to count, add and subtract for answering factoid questions. Models have also been tailored for numerical reasoning  \cite{geva-etal-2020-injecting, zhang-etal-2020-language-embeddings}. % counting, addition and subtraction questions .
 %\texttt{Gen}BERT specifically designed specialized architecture and augmentation method to improve language model's numerical skill \citep{geva-etal-2020-injecting}. 
 \citet{Cobbe2021TrainingVT} proposes to use a verification task during pretraining of LMs to improve their ability to solve math word problems. 
Others have studied monotonicity inference \citep{hu-etal-2019-natural,yanaka-etal-2019-neural,yanaka-etal-2020-neural}, and % Recently, 
\citet{Fang2021PartW} recently focused on the two quantifier words {\em part} and {\em whole} in an error analysis for named entity recognition. 

Many NLU benchmarks contain quantifier words, but their influence on performance has not been studied systematically. One exception to this is that generalized quantifiers have been used to generate adversarial examples in the context of numerical reasoning \citep{naik-etal-2018-stress, nie-etal-2020-adversarial}. 
TaxiNLI \citep{joshi-etal-2020-taxinli}, which categorizes 15 types of reasoning abilities, is a dataset drawn from MNLI. In their taxonomy, the Quantifier category only refers to universal and existential quantifiers, {\em not} to generalized quantifiers, and ditto for \citet{kim-etal-2019-probing}. All of the above focused on English, but in an extension to TaxiNLI, \citet{k-etal-2021-analyzing} incorporated quantifiers into the Logic class and found a large cross-lingual transfer gap on LMs. 

\section{Conclusion}
Quantifiers lie in the intersection of logic, linguistics and NLP research. It is essential for NLU systems to learn quantifier reasoning. We examined generalized quantifiers in multilingual NLU tasks with regards to their expressiveness and logical reasoning requirement. Our survey and experiments indicate quantifiers are neglected to a degree and cause significant performance drops for neural LMs. To better understand LMs' reasoning abilities, we release GQNLI, a novel generalized quantifier NLI challenge dataset. With the pervasiveness of generalized quantifiers, we stress that more efforts are necessary to investigate: (1) when and why models systematically fail when quantifiers interact with other operators; (2) how to improve cross-lingual transferability of quantifiers; (3) how we can exploit the theoretical results about generalized quantifiers from logic and linguistic studies, so as to improve the logical inference ability of neural LMs.

\section*{Acknowledgements}
We would like to thank Miryam de Lhoneux,  Constanza Fierro,  Desmond Elliott and the anonymous reviewers for
their valuable feedback.

\bibliography{anthology,custom}
\bibliographystyle{acl_natbib}

\clearpage
\onecolumn
\appendix
\section*{Appendices}
\label{sec:appendix}

\section{Regular Expressions for Generalized Quantifiers}
\label{app: Regular Expressions}
\input{tables/regular_expressions.tex}
Table~\ref{tab:regex} lists the regex we use to parse generalized quntifiers in sentences augmented with universal dependency tags. The approach does not find all the generalized quantifiers exhuastively but rather approximates the common distributions.

\begin{figure*}[!t]
    \centering
    \includegraphics[width=\textwidth]{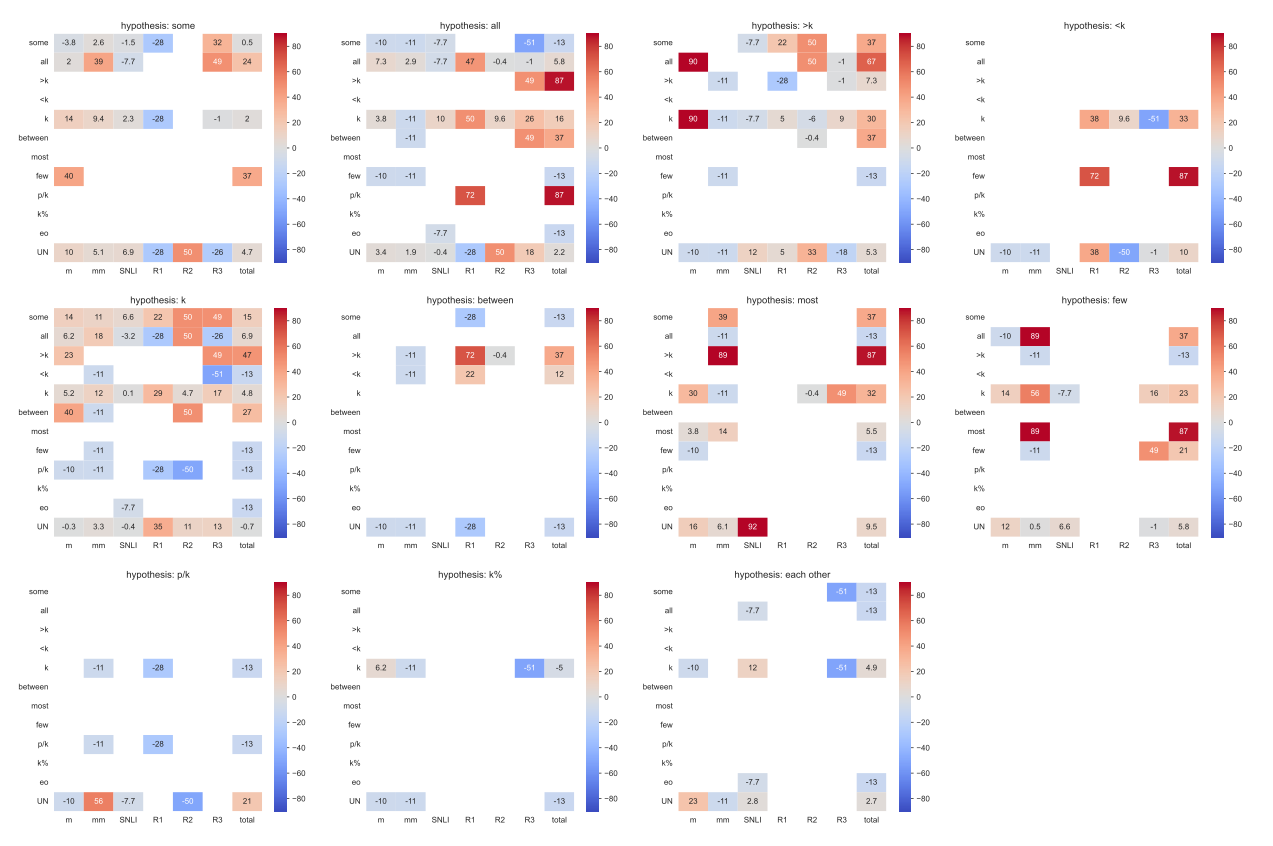}
    \caption{Fine-grained analysis of RoBERTa performance on 6 English NLI subtasks. Each heatmap represents hypotheses with a type of quantifier. The rows stand for premises with the quantifier of that label. The numbers are calculated as the accuracy over the whole dataset minus the fine-grained accuracy given a specific premise and hypothesis (the higher the number, the worse the performance). For each heatmap, the last column represents the accuracy gap weighted by all 6 tasks. ``UN'' stands for an entry where no explicit quantifier is identified.}
    \label{fig:fine_grained}
\end{figure*}

\section{Pairwise Observation}
\label{app:pairwise}
While the analysis in Section \ref{sec:Quantifiers in Four English NLI Tasks} is based on quantifiers in hypotheses, next we consider the interaction of quantifiers in hypotheses and quantifiers in premises. To this end, we calculate the difference between overall performance and performance for premise-hypothesis pairs of GQs. In Figure~\ref{fig:fine_grained}, we visualize the results as heatmaps (see Table~\ref{tab:pairwise} for exact numbers of occurences and accuracies). Surprisingly, whenever  quantifiers appear in both the premise and the hypothesis, LMs largely fail to predict the entailment.  Percentage quantifiers, supposed to be semantically more complex than counting quantifiers, are not {\em de facto} harder in NLI. We studied all 27 cases of percentage quantifiers in the English NLI datasets, and found that in most cases, percentage quantifiers occurrences are {\em identical} across premises and hypotheses, i.e., triggering little or no inference. The other two proportional quantifiers, {\em most} and {\em few}, are hard for LMs to resolve, e.g., in some quantifier pairs, models yield 0\% accuracy. Although {\em each other} is supposed to be hardest to resolve due to the complex semantics of reciprocals \cite{szymanik-thorne-2015-semantic}, it is not reflected in NLI tasks as such. The reason is similar to percentage quantifiers, while annotators intend to alter counting quantifiers when writing hypotheses, reciprocality is seldomly considered a linguistic ability that needs testing for NLU systems. And the annotation for Ramsey quantifier is simply a knockoff, making reciprocal relation identification unwarranted through shallow correlations.
\section{Fine-grained NLI Analysis}

\input{tables/fine-grained_nli_result.tex}

\section{XQA Result: mBERT and XLM-R}
\label{app:xqa_result}
\input{tables/xqa_result_xmlr}
Table~\ref{tab:xqa_both_results} compares the results of mBERT and XLM-R on two XQA tasks, XQuAD and MLQA.

\section{GQNLI Examples}
\label{app:gqnli_example}
Table \ref{tab:gqnli_example} list one example per category in GQNLI.
\input{tables/gqnli_examples}

\section{GQNLI Negation Cases}
\label{app:negation}
We present the results of seven models' performance on cases with negation cues in GQNLI in Table \ref{tab:gqnli_neg}.
\input{tables/x_negation}
\section{GQNLI Subsumption Cases}
\label{app:Subsumption}
See Table \ref{tab:gqnli_sub} for models 'performance on cases requiring subsumption reasoning in GQNLI. We also break down subsumption results by entailment labels into two categories: neutral and non-neutral. 
\input{tables/subsumption_result}

\section{GQNLI Experiment Details}
\label{app:gqnli_hugging}
We reused the fine-tuned BERT and RobERTa in Section \ref{sec:Quantifiers in Four English NLI Tasks}. The other  fine-tuned LMs are from Hugging Face. We list the models and thier links in Table \ref{tab:huggingface}.
\input{tables/huggingface}
\end{document}

%% file: tables/examples.tex
\begin{table}[t]
\centering \footnotesize
\begin{tabular}{p{0.005\columnwidth}p{0.9\columnwidth}}

 \toprule
 \parbox[t]{1mm}{\multirow{2}{*}{\rotatebox[origin=c]{90}{{\textbf{QA\_English}}}}} &
\textsc{Context:} \textit{A piece of paper was later found on which he had written his last statements in \textbf{two} languages, Latin and German. Only \textbf{one} statement was in Latin and the rest in German.} \\ & \textsc{Question:} \textit{In what language were \textbf{most} statements written?} \quad \textsc{Answer:} \texttt{German} \quad \textsc{Predicted Answer:} \texttt{Latin and German}\\

\midrule
\parbox[t]{1mm}{\multirow{2}{*}{\rotatebox[origin=c]{90}{{\textbf{NLI\_Spanish}}}}} &
\textsc{Premise:} \textit{\textbf{Más de tres} personas resultaron heridas en un accidente de \textbf{dos} vehículos el lunes por la noche. (translation: \textbf{More than three} people were injured in a \textbf{two}-vehicle crash  Monday evening.)} \\ & \textsc{Hypothesis:} \textit{Había \textbf{4} personas involucradas. (translation: There were \textbf{4} people involved.} \quad \textsc{Label:} \texttt{Neutral} \quad \textsc{Predicted Label:} \texttt{Entailment}\\
\bottomrule
\end{tabular}
\caption{Examples of quantifiers (marked in bold texts) in NLP tasks, with RoBERTa's prediction for QA and XLM-R's prediction for NLI after fine-tuning.}
\label{tab:examples}
\end{table}

%% file: tables/categorization_set.tex
\begin{table*}[t]
\footnotesize

\centering %\small 
% \scriptsize
\small 
\begin{tabular}{lp{3cm}p{7.6cm}}

\toprule

\textbf{Generalized Quantifiers} & \textbf{Logical Denotation} & \textbf{Example} \\
\midrule
\textbf{some}(A)(B) = 1 & \(A \cap B \neq \varnothing\) & This process is known to increase security in several ways. \\
\textbf{all}(A)(B) = 1 &  \(A \subseteq B\) & Everyone agreed the food was terrible.\\
\textbf{more than k} the(A)(B) = 1 &  \(|A \cap B|>k\) & They do let them go more than twice a week. \\
\textbf{less than k} the(A)(B) = 1 &  \(|A \cap B|<k\) & San Augustin Acolman has less than 1,000 residents.\\
\textbf{k} (A)(B) = 1 &  \(|A \cap B|=k\) & Please donate 100 million to the School of Nursing. \\
\textbf{between p and k} the(A)(B) = 1 &  \(p<|A \cap B|<k\) & The USA added ten states to its nation between 1800 and 1850. \\
the \textbf{p/k} (A)(B) = 1 &  \(|A \cap B|=p \cdot(|A| / k)\) & Captain Blood has 20/20 vision.\\
the \textbf{k\%} (A)(B) = 1 &  \(|A \cap B|=k \cdot(|A| / 100)\) & The lending fund is always guaranteed 9\% interest.\\
\textbf{most} (A)(B) = 1 &  \(|A \cap B|>|A \backslash B|\) & Most ZIP Codes cover roughly ten thousand addresses. \\
\textbf{few} (A)(B) = 1 &  \(|A \cap B|<|A \backslash B|\) & Only a few teenagers were still listening to Rock 'n' Roll.\\
\textbf{each other} (A)(B) = 1 &  \(\forall a \in (A \cap B) 
\exists b \in (A \cap B) (a \neq b)\) & All of these trails are located within the a one hour drive of each other.\\

\bottomrule
\end{tabular}
\caption{The categorization set of quantifiers for task analysis. The first six are Aristotelian/counting quantifiers and the following four are proportional quantifiers. The last one is a Ramsey quantifier \citep{Schmerl1982OnTR}. For each quantifier, its logical denotation is listed in the second column. The third conlumn contains English examples with quantifiers taken from XNLI.}
\label{tab:categorization_set}
\end{table*}

%% file: tables/task_stats_nli.tex
\begin{table}[t]
\centering %\small 
% \scriptsize
% \footnotesize
\resizebox{\columnwidth}{!}{%
\begin{tabular}{lcccccccc}

\toprule

 \multirow{4}{*}{Quantifier} & \multicolumn{6}{c}{English} & \multirow{2}{1cm}{Cross-lingual} \\
  & \rotatebox{90}{MNLI\_m} & \rotatebox{90}{MNLI\_mm} & \rotatebox{90}{SNLI} & \rotatebox{90}{ANLI\_R1} & \rotatebox{90}{ANLI\_R2} & \rotatebox{90}{ANLI\_R3} & \rotatebox{90}{XNLI}  \\
 \midrule
 some & 171 & 132 & 191 & 5 & 1 & 17 & 115  \\
 all & 255 & 239 & 65 & 15 & 8 & 29 & 166  \\
\(>k\) & 14 & 23 & 8 & 10 & 16 & 14 & 16  \\
\(<k\) & 3 & 3 & 0 & 6 & 7 & 5 & 1  \\
\(k\) & 266 & 269 & 988 & 55 & 62 & 48 & 159 \\
between & 2 & 3 & 0 & 3 & 2 & 0 & 1 \\
\(p/k\) & 1 & 5 & 1 & 1 & 1 & 0 & 2 \\
\(k \%\) & 10 & 7 & 0 & 0 & 0 & 1 & 5 \\
most & 35 & 39 & 1 & 0 & 2 & 1 & 9\\
 few & 14 & 15 & 11 & 0 & 0 & 6 & 11  \\
 each other & 4 & 3 & 35 & 0 & 0 & 2 & 5  \\
\midrule
Total & 775 & 738 & 1300 & 95 & 99 & 124 & 499  \\
Frequency & 7.9\% & 7.5\% & 13.2\% & 9.5\% & 9.9\% & 12.4\% & 10.0\% \\

\bottomrule
\end{tabular}
}
\caption{Quantifier distribution in four NLI tasks, among which three are monolingual English and one is cross-lingual. The table show statistics of the test set, if not available, dev set, of the target task. All but the last rows show the occurrence time of the type of quantifier in the first column. The last row represents the distribution rate of any quantifier in the dataset.  }
\label{tab:task_stats_nli}
\end{table}

%% file: tables/nli_result.tex
\begin{table*}[t]
\centering 
% \small 
\scriptsize
% \footnotesize
\resizebox{2\columnwidth}{!}{%
\begin{tabular}{lccccccc|ccccccc}
\toprule
 \multirow{4}{*}{Quantifier} & \multicolumn{7}{c|}{BERT} & \multicolumn{7}{c}{RoBERTa}\\
      & \rotatebox{90}{M\_m} & \rotatebox{90}{M\_mm} & \rotatebox{90}{SNLI} & \rotatebox{90}{A\_R1} & \rotatebox{90}{A\_R2} & \rotatebox{90}{A\_R3} & \rotatebox{90}{\textit{weig.}} & \rotatebox{90}{M\_m} & \rotatebox{90}{M\_mm} & \rotatebox{90}{SNLI} & \rotatebox{90}{A\_R1} & \rotatebox{90}{A\_R2} & \rotatebox{90}{A\_R3} & \rotatebox{90}{\textit{weig.}} \\
 \midrule
 some & \textbf{82.5} & \textbf{84.1} & \textbf{86.9} & 100 & \textbf{0} & 47.1 & \textbf{83.4} & \textbf{83} & \textbf{84.8} & \textbf{86.9} & 100 & 100 & \textbf{41.1} & \textbf{83.7}\\
 all & \textbf{85.9} & 88.3 & \textbf{89.2} & \textbf{46.7} & \textbf{37.5} & \textbf{34.5} & \textbf{83.2} & \textbf{85.9} & 92.1 & \textbf{90.8} & \textbf{66.7} & \textbf{37.5} & \textbf{34.5} & \textbf{85.3}\\
\(>k\) & \textbf{85.7} & 100 & \textbf{87.5} & 70 & \textbf{43.8} & \textbf{42.9} & \textbf{73} & \textbf{85.7} & 91.3 & \textbf{87.5} & 80 & \textbf{37.5} & \textbf{28.5} & \textbf{68.2} \\
\(<k\) & 100 & 100 & \/ & \textbf{33.3} & 57.1 & 80 & \textbf{66.7} & 100 & 100 & \/ & 83.3 & 85.7 & 100 & 91.7  \\
\(k\) & 87.2 & \textbf{81.8} & 92.4 & \textbf{43.6} & \textbf{43.5} & \textbf{33.3} & 84.8 & 88.3 & \textbf{88.8} & 92.9 & \textbf{56.3} & 61.3 & \textbf{43.8} & 87.8  \\
between & 100 & 100 & \/ & 66.7 & 50 & \/ & \textbf{80} & 100 & \textbf{66.7} & \/ & \textbf{66.7} & 50 & \/ & \textbf{70} \\
\(p/k\) & 100 & \textbf{60} & 100 & 100 & 100 & \/ & \textbf{77.8} & 100 & \textbf{80} & 100 & 100 & \textbf{0} & \/ & \textbf{77.8} \\
\(k \%\) & 90 & 100 & \/ & \/ & \/ & 100 & 94.4 & \textbf{70} & \textbf{85.7} & \/ & \/ & \/ & \textbf{0} & \textbf{72.2} \\
most & \textbf{74.3} & \textbf{79.5} & \textbf{0} & \/ & 50 & \textbf{0} & \textbf{74.4} & \textbf{77} & \textbf{87.2} & 100 & \/ & 59 & \textbf{0} & \textbf{80.9}\\
 few & \textbf{78.6} & \textbf{73.3} & \textbf{90.9} & \/ & \/ & \textbf{33.3} & \textbf{73.9} & \textbf{85.7} & \textbf{80} & \textbf{90.9} & \/ & \/ & \textbf{33.3} & \textbf{78.3}  \\
 each other & \textbf{75} & 100 & \textbf{85.7} & \/ & \/ & 50 & \textbf{84.1} & \textbf{50} & 100 & \textbf{88.6} & \/ & \/ & 50 & \textbf{84.1}  \\
\midrule
all GQs & \textbf{85} & \textbf{84.8} & \textbf{91.2} & \textbf{50.5} & \textbf{44.4} & \textbf{39} & \textbf{83.3} & \textbf{85.4} & \textbf{88.8} & \textbf{91.7} & \textbf{65.3} & 56.5 & \textbf{40.3} & \textbf{85.5}  \\
full & 86.5 & 86.1 & 91.3 & 58.6 & 48 & 43.2 & 84.4 & 89.5 & 89.4 & 92.3 & 71.7 & 49.6 & 49 & 87.3 \\

\bottomrule
\end{tabular}
}
\caption{BERT and RoBERTa performance on NLI tasks. The \textit{weig.} column represents the percentage of all true predictions in six subtasks over total instances. The penultimate row stands for the overall performance when quantifiers exist in a dataset. The last row reports the overall performance in a dataset. Number marked in bold signifies a lower score than the overall performance.}
\label{tab:nli_result}
\end{table*}

%% file: tables/xnli_result.tex
\begin{table*}[t]
\centering %\small 
% \scriptsize
% \footnotesize
\resizebox{2\columnwidth}{!}{%
\begin{tabular}{lccccccc|ccccccc}
\toprule
 \multirow{2}{*}{Quantifier} & \multicolumn{7}{c|}{mBERT} & \multicolumn{7}{c}{XLM} \\
  & en & zh & es & ar & vi & de & \textit{weig.} & en & zh & es & ar & vi & de & \textit{weig.}  \\
 \midrule
some & 85.2 & \textbf{69.6} & 80 & \textbf{63.5} & \textbf{67.8} & 74.8 & 73.4 & 85.2 & 70.3 & 79.1 & 71.3 & 73.9 & \textbf{69.6} & \textbf{69.6}\\
all & \textbf{80.1} & \textbf{65.7} & \textbf{72.8} & 69.3 & \textbf{63.9} & 74.1 & \textbf{70.9} & \textbf{82.5} & \textbf{62.7} & 74.1 & 67.5 & 71.7 & 73.5 & \textbf{72} \\
\(>k\) & 87.5 & \textbf{50} & \textbf{68.8} & \textbf{43.8} & \textbf{56.2} & \textbf{62.5} & \textbf{61.6} & \textbf{81.2} & \textbf{62.5} & \textbf{56.2} & \textbf{62.5} & \textbf{50} & 75 & 75\\
\(<k\) & 100 & 100 & 100 & 100 & 100 & 100 & 100 & 100 & 100 & 100 & 100 & 100 & 100 & 100\\
\(k\) & 86.2 & \textbf{69.1} & 80.5 & 71.7 & 76.7 & 82.4 & 77.7 & 83 & 66.7 & 78.6 & 71.7 & 74.2 & 81.1 & 75.8\\
between & 100 & 100 & 100 & 100 & 100 & 100 & 100 & 100 & 100 & 100 & 100 & 100 & 100 & 100\\
\(p/k\) & 100 & \textbf{50} & 100 & 100 & 100 & 100 & 91.7 & 100 & \textbf{0} & 100 & 100 & \textbf{50} & \textbf{50} & \textbf{66.7} \\
\(k \%\) & 100 & 100 & 80 & 100 & 100 & 100 & 96.7 & \textbf{80} & 80 & 80 & 100 & 100 & 80 & 86.7\\
most & \textbf{55.6} & \textbf{55.6} & \textbf{66.7} & 66.7 & \textbf{33.3} & 66.7 & \textbf{57.4} & \textbf{55.6} & \textbf{33.3} & \textbf{66.7} & \textbf{55.6} & \textbf{44.4} & 77.8 & \textbf{55.6}\\
few & \textbf{72.7} & \textbf{54.5} & \textbf{72.7} & \textbf{63.6} & \textbf{45.5} & 72.7 & \textbf{63.6} & \textbf{63.6} & \textbf{36.4} & \textbf{54.5} & \textbf{63.6} & \textbf{54.5} & 72.7 & \textbf{57.5}\\
each other & \textbf{60} & \textbf{60} & \textbf{60} & \textbf{60} & 80 & 80 & \textbf{66.7} & \textbf{80} & \textbf{20} & \textbf{60} & \textbf{20} & \textbf{40} & \textbf{60} & \textbf{46.7}\\
\midrule
all GQs & 83 & \textbf{67.1} & 76.7 & 68.1 & \textbf{68.3} & 76.9 & 73.3 & \textbf{82.4} & \textbf{64.2} & \textbf{75.7} & 69.3 & \textbf{71.4} & 74.8 & 73\\
comp. & 82.6 & 88.9 & 74.7 & 65.6 & 70.7 & 71.4 & 72.4 & 83.1 & 64.8 & 76.3 & 66.9 & 71.6 & 71.3 & 72.3 \\

\bottomrule
\end{tabular}
}
\caption{Results of mBERT and XLM performance on XNLI tasks decomposed by quantifier categories.}
\label{tab:xnli_result}
\end{table*}

%% file: tables/task_stats_qa.tex
\begin{table}[t]
\centering %\small 
% \scriptsize
% \footnotesize
\resizebox{\columnwidth}{!}{%
\begin{tabular}{lccccccc}

\toprule

 \multirow{2}{*}{Quantifier} & \multicolumn{5}{c}{MLQA} & \multicolumn{2}{r}{XQuAD} \\
  & en & zh & es & ar & vi & de & ...  \\
 \midrule
 some & 66 & 39 & 41 & 44 & 37 & 33 & 12   \\
 all & 31 & 14 & 26 & 21 & 19 & 16 &  7  \\
\(<k\) & 1 & 0 & 0 & 0 & 1 & 0 &  0  \\
\(k\) & 322 & 168 & 166 & 195 & 204 & 149 & 32   \\
between & 4 & 2 & 2 & 2 & 3 & 0 & 3  \\
\(p/k\) & 1 & 1 & 1 & 0 & 0 & 0 &  0 \\
\(k \%\) & 1 & 1 & 0 & 1 & 0 & 0 & 0  \\
most & 27 & 19 & 11 & 30 & 17 & 9 &  5 \\
\midrule
Total & 453 & 244 & 247 & 293 & 281 & 207 & 59    \\
Frequency & 3.9\% & 4.7\% & 4.7\% & 5.4\% & 5.1\% & 4.5\% & 5.0\%  \\

\bottomrule
\end{tabular}
}
\caption{Quantifier distribution in two multilingual QA tasks, MLQA and XQuAD. We choose six common languages apprearing in both tasks to facilitate comparisons. XQuAD is strictly parellel while MLQA is not, hence only the latter has statistics by languges. Categories that no entry exists are omitted.}
\label{tab:task_stats_qa}
\end{table}

%% file: tables/xqa_result_mbert.tex
\begin{table*}[t]
\centering %\small 
% \scriptsize
% \footnotesize
\resizebox{\textwidth}{!}{%
\begin{tabular}{lcccccccccccccc|cccccccccccccc}

\toprule

 \multirow{3}{*}{Quantifier} & \multicolumn{14}{c|}{XQuAD} & \multicolumn{14}{c}{MLQA} \\
  & \multicolumn{2}{c}{en} & \multicolumn{2}{c}{zh} & \multicolumn{2}{c}{es} & \multicolumn{2}{c}{ar} & \multicolumn{2}{c}{vi} & \multicolumn{2}{c}{de} & \multicolumn{2}{c|}{\textit{weighted}} & \multicolumn{2}{c}{en} & \multicolumn{2}{c}{zh} & \multicolumn{2}{c}{es} & \multicolumn{2}{c}{ar} & \multicolumn{2}{c}{vi} & \multicolumn{2}{c}{de} & \multicolumn{2}{c}{\textit{weighted}}\\
 & EM & F1 & EM & F1 & EM & F1 & EM & F1 & EM & F1 & EM & F1 & EM & F1 & EM & F1 & EM & F1 & EM & F1 & EM & F1 & EM & F1 & EM & F1 & EM & F1 \\
  \midrule
some &75 & 84.2 & 50 & \textbf{55.5} & 58.3 & 76.1 & 50 & \textbf{50} & 16.6 & 42.4 & \textbf{33.3} & \textbf{43.8} & 47.2 & 58.7 & \textbf{59} & 80 & \textbf{28.2} & \textbf{52.1} & \textbf{34.1} & \textbf{59.2} & 36.3 & 54.9 & \textbf{5.4} & \textbf{24} & \textbf{33.3} & 58.4 & \textbf{32.7} & \textbf{54.8}\\
all & \textbf{28.5} & \textbf{62.2} & \textbf{14.2} & \textbf{35.2} & \textbf{28.5} & 82 & 42.8 & \textbf{52.3} & \textbf{14.2} & \textbf{29.4} & \textbf{28.5} & \textbf{56} & \textbf{26.1} & \textbf{52.9} &  67.7 & \textbf{79.8} & 14.2 & \textbf{46.4} & \textbf{38.4} & \textbf{62.8} & 33.3 & 57.9 & \textbf{10.5} & 30.1 &\textbf{31.2} & \textbf{51.6} & \textbf{32.6} & \textbf{54.8} \\
\(<k\) & &  &  &  &  &  &  &  &  &  &  &  &  &  & \textbf{0} & \textbf{0} &  &  &  &  &  &  & 0 & \textbf{13.3} &  &  & \textbf{0} & \textbf{6.7}     \\
\(k\) &78.1 & 90.1 & 68.7 & 80.4 & 56.2 & \textbf{72.1} & \textbf{40.6} & 64.3 & \textbf{12.5} & 35.7 & 56.2 & 77.1 & 52.1 & 70 & 74.9 & \textbf{79.4} & 47 & 63.4 & \textbf{41.5} & \textbf{65.9} & \textbf{27.6} & 50.3 & \textbf{6.3} & \textbf{23.7} & \textbf{38.2} & \textbf{53} & 39.3 & \textbf{56} \\
between & 100 & 100 & \textbf{33.3} & \textbf{72.2} & 66.6 & 93.3 & 100 & 100 & \textbf{0} & \textbf{19} & \textbf{0} & \textbf{56.5} & 50 & 73.5 & \textbf{50} & 88.5 & 50 & 83.3 & \textbf{0} & \textbf{26.6} & \textbf{0} & 68.7 & \textbf{0} & \textbf{26.6} &  &  & \textbf{20} & 58.7 \\
\(p/k\) &  &  &  &  &  &  &  &  &  &  &  &  & & & 100 & 100& \textbf{0} & \textbf{0} & \textbf{0} & \textbf{0} &  &  &  &  &  &  & \textbf{33.3} & \textbf{33.3}   \\
\(k \%\) &  &  &  &  &  &  &  &  &  &  &  & & & & 100 & 100 & \textbf{0} & \textbf{26.6} &  &  & \textbf{0} & \textbf{23.7} &  &  &  &  & \textbf{33.3} & \textbf{50.1}   \\
most & \textbf{40} & \textbf{53.3} & \textbf{40} & \textbf{40} & \textbf{0} & \textbf{10} & \textbf{0} & \textbf{26.6} & \textbf{0} & \textbf{0} & \textbf{20} & \textbf{49.3} & \textbf{16.7} & \textbf{29.9} & \textbf{55.5} & \textbf{76} & 47.3 & 62.1 & \textbf{45.4} & \textbf{61.7} & 30 & \textbf{46.8} & \textbf{5.8} & \textbf{15.7} & \textbf{33.3} & \textbf{40.7} & \textbf{36.2} & \textbf{50.3} \\
\midrule
all GQs & \textbf{70} & \textbf{83.2} & 55 & 66.7 & \textbf{50} & \textbf{70.3} & 41.6 & 58.2 & \textbf{11.6} & 32.5 & \textbf{43.3} & \textbf{65} & \textbf{45.3} & 62.7 & \textbf{63.5} & \textbf{79.2} & 41.8 & 60.3 & \textbf{39.6} & \textbf{63.7} & \textbf{29.3} & 51.3 & \textbf{6.4} & \textbf{23.6} & \textbf{36.1} & \textbf{53.2} & \textbf{36.1} & \textbf{55.2} \\
comp. & 71.8 & 83.7 & 48 & 59.1 & 56 & 74.5 & 40.8 & 57.9 & 13.9 & 32.4 & 50.7 & 67.2 & 46.9 & 62.5 &  67.2 & 80.6 & 37.5 & 57.9 & 47.3 & 66 & 30 & 48.4 & 11.2 & 28 & 40.8 & 56 & 39 & 56.2\\
\bottomrule

\end{tabular}
}
\caption{Results of mBERT performance on XQA tasks decomposed by quantifier categories.}
\label{tab:xqa_result_mbert}
\end{table*}

%% file: tables/gqnli.tex
\begin{table*}[t]
    \small
    \resizebox{\textwidth}{!}{%
    \begin{tabular}{llccccccccccc|c}
    \toprule
    
    \multicolumn{2}{c}{\textbf{Quantifier}} & some & all & \(>k\) & \(<k\) & k & between & \(p/k\) & \(k \%\) & most & few & each other & \textit{Overall} \\ 
     \multicolumn{2}{c}{\textbf{\# Occurrence}} & 27 & 51 & 51 & 33 & 170 & 21 & 24 & 45 & 18 & 9 & 36 & 485 \\ \midrule
    \textbf{Model} & \textbf{Training Data} & \multicolumn{12}{c}{\textbf{\% Performance}} \\ \midrule
\multirow{1}{*}{BERT} & S,M,F,ANLI & 40.7 & 41.2 & 33.3 & 30.3 & 30.6 & 14.3 & 37.5 & 22.2 & 61.1 & 22.2 & 41.7 & 30 \\ \midrule
 \multirow{1}{*}{ELECTRA} & S,M,F,ANLI & 37.0 & 17.6 & 54.9 & 27.3 & 38.2 & 14.3 & 62.5 & 31.1 & 61.1 & 0.0 & 16.7 & 38.0 \\ \midrule
 \multirow{1}{*}{SBERT} & S,M,F,ANLI & 66.7 & 43.1 & 47.1 & 24.2 & 32.4 & 14.3 & 25.0 & 31.1 & \textbf{77.8} & \textbf{66.7} & 36.1 & 39.3 \\ \midrule
 \multirow{2}{*}{RoBERTa} & MNLI & 55.6 & 25.5 & 17.6 & 27.3 & 24.7 & 23.8 & 45.8 & 17.8 & 33.3 & 33.3 & 11.1 & 28.2\\ 
     & S,M,F,ANLI & 63.0 & 41.2 & 41.2 & 27.3 & 34.1 & 28.6 & \textbf{75.0} & 33.3 & 50.0 & 33.3 & 38.9 & 39.3 \\ \midrule
\multirow{1}{*}{ALBERT} & S,M,F,ANLI & 70.4 & 45.1 & 35.3 & 33.3 & 36.5 & 19.0 & 37.5 & 37.8 & 50.0 & 11.1 & 36.1 & 41.7 \\ \midrule
 \multirow{2}{*}{BART} & MNLI & 40.7 & 21.6 & \textbf{60.8} & 36.4 & \textbf{50.6} & \textbf{66.7} & 37.5 & 46.7 & 27.8 & 33.3 & 22.2 & 41.3 \\ 
     & S,M,F,ANLI & 59.3 & 51.0 & 35.3 & 30.3 & 35.3 & 19.0 & 66.7 & 20.0 & 50.0 & \textbf{66.7} & 47.2 & 42.7 \\ \midrule
\multirow{3}{*}{DeBERTa-v3} & MNLI & 48.1 & 37.3 & 33.3 & 33.3 & 35.9 & 33.3 & 41.7 & 33.3 & 33.3 & 33.3 & 41.7 & 34.7 \\ 
     & M,F,ANLI & \textbf{81.5} & 54.9 & 49.0 & 33.3 & 44.7 & 28.6 & 50.0 & \textbf{48.9} & 66.7 & 55.6 & 44.4 & \textbf{48.0} \\ 
     & M,F,Ling,DocNLI & 77.8 & \textbf{70.6} & 49.0 & \textbf{54.5} & 44.7 & 4.8 & 33.3 & 42.2 & 50.0 & \textbf{66.7} & \textbf{58.3} & 45.0\\ \bottomrule
     \end{tabular}
     }
\caption{GQNLI statstics and seven types of models' performance with different combinations of training data. The second row shows the occurrence time of the type of GQ in GQNLI. The following rows show models' performance on the dataset. We tested most competitive models fine-tuned for NLI available on Hugging Face. All but ALBERT (\texttt{xxlarge}) and DeBERTa-v3 (\texttt{base}) are size \texttt{large}. S, M, F, Ling, A, DocNLI refer to SNLI, MNLI, Fever-NLI, LingNLI \citep{parrish-etal-2021-putting-linguist}, ANLI and DocNLI \citep{yin-etal-2021-docnli}, respectively. Numbers in bold represent the highest accuracy in one category. Due to space limitation we provide the link to each model in the Appendix \ref{app:gqnli_hugging}.}
\label{tab:gqnli}
\end{table*}

%% file: tables/regular_expressions.tex
\begin{table*}[t]
\footnotesize

\centering %\small 
\scriptsize
% \small 
\begin{tabular}{lp{12cm}p{4cm}}

\toprule

\textbf{Generalized Quantifiers} & \textbf{Regular Expressions} \\
\midrule
\textbf{some}(A)(B) = 1 & \begin{lstlisting}[breaklines]
(some|several|much|many)\/det .*\/(nsubj|obj|obl)|(some|several|much|many)\/nsubj|(some|several|much|many)\/amod \w+\/nsubj:pass\end{lstlisting}
 \\
\textbf{all}(A)(B) = 1 & \begin{lstlisting}[breaklines]
(every|all|each)\/det .*\/(nsubj|obj|obl)|all\/det:predet .*\/(nsubj|obj|obl) |everything|everyone|everybody
\end{lstlisting}\\
\textbf{more than k} the(A)(B) = 1 &  \begin{lstlisting}[breaklines]
((more|great)\/advmod than\/(fixed|case)|at\/case least\/nmod) .+\/nummod .+\/(nsubj|obj|obl)\end{lstlisting} \\
\textbf{less than k} the(A)(B) = 1 & \begin{lstlisting}[breaklines]
((few|less)\/advmod than\/(fixed|case)|at\/case most\/amod) .+\/nummod .+\/(nsubj|obj|obl) \end{lstlisting}\\
\textbf{k} (A)(B) = 1 & \begin{lstlisting}[breaklines]
\w+\/nummod .+\/(nsubj|obj|obl)\end{lstlisting} \\
\textbf{between p and k} the(A)(B) = 1 & \begin{lstlisting}[breaklines]
between\/case \w+\/(nummod|nsubj|obj|obl) and\/cc \w+\/conj|between\/case .+\/(nummod|nsubj|obj|obl) %\/obl\end{lstlisting} \\
the \textbf{p/k} (A)(B) = 1 & \begin{lstlisting}[breaklines]
\d+\/\d+\/(nummod|nsubj|obj|obl)|half\/nummod|third\/(nsubj|obj|obl)|fourth\/(nsubj|obj|obl)|fifth\/(nsubj|obj|obl)\end{lstlisting} \\
the \textbf{k\%} (A)(B = 1 ) &  \begin{lstlisting}[breaklines]
\d+\/nummod %\/(nsubj|obj|obl)\end{lstlisting} \\
\textbf{most} (A)(B) = 1 &  \begin{lstlisting}[breaklines]
most\/amod \w+\/(nsubj|obj|obl)|most\/nsubj:pass of\/case .+\/nmod \end{lstlisting}\\
\textbf{few} (A)(B) = 1 & \begin{lstlisting}[breaklines]
few\/amod \w+\/(nsubj|obj|obl)|few\/nsubj:pass of\/case .+\/nmod \end{lstlisting}\\
\textbf{each other} (A)(B) = 1 &  \begin{lstlisting}[breaklines]
each\/det other\/(nsubj|obj|obl)\end{lstlisting} \\

\bottomrule
\end{tabular}

\caption{Regular Expressions for generalized quantifiers.}
\label{tab:regex}
\end{table*}

%% file: tables/fine-grained_nli_result.tex
\begin{table*}[t]
\scriptsize
\resizebox{\textwidth}{!}{%
\begin{tabular}{lllllllllllllllll}
 \multirow{2}{*}{Hypothesis} & \multirow{2}{*}{Premise} & MNLI\_m\_dev &  & MNLI\_mm\_dev &  & SNLI\_test &  & ANLI\_R1\_test &  & ANLI\_R2\_test &  & ANLI\_R3\_test &  & Total &  &  \\
 &  & \#occurrence & \%Acc & \#occurrence & \%Acc & \#occurrence & \%Acc & \#occurrence & \%Acc & \#occurrence & \%Acc & \#occurrence & \%Acc & \#occurrence & \%Acc & \#correctpred \\
some & some & 45 & 93.3 & 38 & 86.8 & 16 & 93.8 & 1 & 100 & \/ & \/ & 6 & 16.7 & 106 & 86.8 & 92 \\
 & all & 8 & 87.5 & 8 & 50 & 3 & 100 & \/ & \/ & \/ & \/ & 3 & 0 & 22 & 63.6 & 14 \\
 & \textgreater{}k & \/ & \/ & \/ & \/ & \/ & \/ & \/ & \/ & \/ & \/ & \/ & \/ & \/ & \/ & \/ \\
 & \textless{}k & \/ & \/ & \/ & \/ & \/ & \/ & \/ & \/ & \/ & \/ & \/ & \/ & \/ & \/ & \/ \\
 & k & 12 & 75 & 10 & 80 & 40 & 90 & 4 & 100 & 0 &  & 2 & 50 & 68 & 85.3 & 58 \\
 & between & \/ & \/ & \/ & \/ & \/ & \/ & \/ & \/ & \/ & \/ & \/ & \/ & \/ & \/ & \/ \\
 & most & \/ & \/ & \/ & \/ & \/ & \/ & \/ & \/ & \/ & \/ & \/ & \/ & \/ & \/ & \/ \\
 & few & 2 & 50 & \/ & \/ & \/ & \/ & \/ & \/ & \/ & \/ & \/ & \/ & 2 & 50 & 1 \\
 & p/k & \/ & \/ & \/ & \/ & \/ & \/ & \/ & \/ & \/ & \/ & \/ & \/ & \/ & \/ & \/ \\
 & k\% & \/ & \/ & \/ & \/ & \/ & \/ & \/ & \/ & \/ & \/ & \/ & \/ & \/ & \/ & \/ \\
 & eachother & \/ & \/ & \/ & \/ & \/ & \/ & \/ & \/ & \/ & \/ & \/ & \/ & \/ & \/ & \/ \\
 & "unmatched" & 110 & 79.1 & 83 & 84.3 & 137 & 85.4 & 1 & 100 & 1 & 0 & 8 & 75 & 340 & 82.6 & 281 \\
 &  & \/ & \/ & \/ & \/ & \/ & \/ & \/ & \/ & \/ & \/ & \/ & \/ & \/ & \/ & \/ \\
all & some & 11 & 100 & 12 & 100 & 2 & 100 & \/ & \/ & \/ & \/ & 1 & 100 & 26 & 100 & 26 \\
 & all & 73 & 82.2 & 74 & 86.5 & 3 & 100 & 4 & 25 & 2 & 50 & 6 & 50 & 162 & 81.5 & 132 \\
 & \textgreater{}k & \/ & \/ & \/ & \/ & \/ & \/ & \/ & \/ & \/ & \/ & 1 & 0 & 1 & 0 & 0 \\
 & \textless{}k & \/ & \/ & \/ & \/ & \/ & \/ & \/ & \/ & \/ & \/ & \/ & \/ & \/ & \/ & \/ \\
 & k & 28 & 85.7 & 19 & 100 & 22 & 81.8 & 9 & 22.2 & 5 & 40 & 13 & 23.1 & 96 & 70.8 & 68 \\
 & between & \/ & \/ & 1 & 100 & \/ & \/ & \/ & \/ & \/ & \/ & 1 & 0 & 2 & 50 & 1 \\
 & most & \/ & \/ & \/ & \/ & \/ & \/ & \/ & \/ & \/ & \/ & \/ & \/ & \/ & \/ & \/ \\
 & few & 4 & 100 & 2 & 100 & \/ & \/ & \/ & \/ & \/ & \/ & \/ & \/ & 6 & 100 & 6 \\
 & p/k & \/ & \/ & \/ & \/ & \/ & \/ & 1 & 0 & \/ & \/ & \/ & \/ & 1 & 0 & \/ \\
 & k\% & \/ & \/ & \/ & \/ & \/ & \/ & \/ & \/ & \/ & \/ & \/ & \/ & \/ & \/ & \/ \\
 & eachother & 0 &  & 0 &  & 1 & 100 & 0 &  & 0 &  & 0 &  & 1 & 100 & 1 \\
 & "unmatched" & 151 & 86.1 & 144 & 87.5 & 41 & 92.7 & 5 & 100 & 2 & 0 & 13 & 30.8 & 356 & 85.1 & 303 \\
 &  & \/ & \/ & \/ & \/ & \/ & \/ & \/ & \/ & \/ & \/ & \/ & \/ & \/ & \/ & \/ \\
\textgreater{}k & some & \/ & \/ & \/ & \/ & 1 & 100 & 2 & 50 & 1 & 0 & \/ & \/ & 4 & 50 & 2 \\
 & all & 1 & 0 & \/ & \/ & \/ & \/ & \/ & \/ & 2 & 0 & 2 & 50 & 5 & 20 & 1 \\
 & \textgreater{}k & \/ & \/ & 2 & 100 & \/ & \/ & 1 & 100 & \/ & \/ & 2 & 50 & 5 & 80 & 4 \\
 & \textless{}k & \/ & \/ & \/ & \/ & \/ & \/ & \/ & \/ & \/ & \/ & \/ & \/ & \/ & \/ & \/ \\
 & k & 1 & 0 & 3 & 100 & 2 & 100 & 3 & 66.7 & 9 & 55.6 & 10 & 40 & 28 & 57.1 & 16 \\
 & between & \/ & \/ & \/ & \/ & \/ & \/ & \/ & \/ & 2 & 50 & \/ & \/ & 2 & 50 & 1 \\
 & most & \/ & \/ & \/ & \/ & \/ & \/ & \/ & \/ & \/ & \/ & \/ & \/ & \/ & \/ & \/ \\
 & few & \/ & \/ & 1 & 100 & \/ & \/ & \/ & \/ & \/ & \/ & \/ & \/ & 1 & 100 & 1 \\
 & p/k & \/ & \/ & \/ & \/ & \/ & \/ & \/ & \/ & \/ & \/ & \/ & \/ & \/ & \/ & \/ \\
 & k\% & \/ & \/ & \/ & \/ & \/ & \/ & \/ & \/ & \/ & \/ & \/ & \/ & \/ & \/ & \/ \\
 & eachother & \/ & \/ & \/ & \/ & \/ & \/ & \/ & \/ & \/ & \/ & \/ & \/ & \/ & \/ & \/ \\
 & "unmatched" & 12 & 100 & 18 & 100 & 5 & 80 & 6 & 66.7 & 6 & 16.7 & 3 & 66.7 & 50 & 82 & 41 \\
 &  & \/ & \/ & \/ & \/ & \/ & \/ & \/ & \/ & \/ & \/ & \/ & \/ & \/ & \/ & \/ \\
\textless{}k & some & \/ & \/ & \/ & \/ & \/ & \/ & \/ & \/ & \/ & \/ & \/ & \/ & \/ & \/ & \/ \\
 & all & \/ & \/ & \/ & \/ & \/ & \/ & \/ & \/ & \/ & \/ & \/ & \/ & \/ & \/ & \/ \\
 & \textgreater{}k & \/ & \/ & \/ & \/ & \/ & \/ & \/ & \/ & \/ & \/ & \/ & \/ & \/ & \/ & \/ \\
 & \textless{}k & \/ & \/ & \/ & \/ & \/ & \/ & \/ & \/ & \/ & \/ & \/ & \/ & \/ & \/ & \/ \\
 & k & \/ & \/ & \/ & \/ & \/ & \/ & 3 & 33.3 & 5 & 40 & 3 & 100 & 11 & 54.5 & 6 \\
 & between & \/ & \/ & \/ & \/ & \/ & \/ & \/ & \/ & \/ & \/ & \/ & \/ & \/ & \/ & \/ \\
 & most & \/ & \/ & \/ & \/ & \/ & \/ & \/ & \/ & \/ & \/ & \/ & \/ & \/ & \/ & \/ \\
 & few & \/ & \/ & \/ & \/ & \/ & \/ & 1 & 0 & \/ & \/ & \/ & \/ & 1 & 0 & 0 \\
 & p/k & \/ & \/ & \/ & \/ & \/ & \/ & \/ & \/ & \/ & \/ & \/ & \/ & \/ & \/ & \/ \\
 & k\% & \/ & \/ & \/ & \/ & \/ & \/ & \/ & \/ & \/ & \/ & \/ & \/ & \/ & \/ & \/ \\
 & eachother & \/ & \/ & \/ & \/ & \/ & \/ & \/ & \/ & \/ & \/ & \/ & \/ & \/ & \/ & \/ \\
 & "unmatched" & 3 & 100 & 3 & 100 & \/ & \/ & 3 & 33.3 & 2 & 100 & 2 & 50 & 13 & 76.9 & 10 \\
 & \/ & \/ & \/ & \/ & \/ & \/ & \/ & \/ & \/ & \/ & \/ & \/ & \/ & \/ & \/ & \/ \\
k & some & 8 & 75 & 14 & 78.6 & 28 & 85.7 & 2 & 50 & 2 & 0 & 4 & 0 & 58 & 72.4 & 42 \\
 & all & 12 & 83.3 & 14 & 71.4 & 22 & 95.5 & 1 & 100 & 3 & 0 & 4 & 75 & 56 & 80.4 & 45 \\
 & \textgreater{}k & 3 & 66.7 & \/ & \/ & \/ & \/ & \/ & \/ & \/ & \/ & 2 & 0 & 5 & 40 & 2 \\
 & \textless{}k & \/ & \/ & 2 & 100 & \/ & \/ & \/ & \/ & \/ & \/ & 1 & 100 & 3 & 100 & 3 \\
 & k & 140 & 84.3 & 121 & 76.9 & 593 & 92.2 & 42 & 42.9 & 49 & 44.9 & 37 & 32.4 & 982 & 82.5 & 810 \\
 & between & 2 & 50 & 2 & 100 & \/ & \/ & \/ & \/ & 1 & 0 & \/ & \/ & 5 & 60 & 3 \\
 & most & \/ & \/ & \/ & \/ & \/ & \/ & \/ & \/ & \/ & \/ & \/ & \/ & \/ & \/ & \/ \\
 & few & \/ & \/ & 1 & 100 & \/ & \/ & \/ & \/ & \/ & \/ & \/ & \/ & 1 & 100 & 1 \\
 & p/k & 1 & 100 & 1 & 100 & \/ & \/ & 1 & 100 & 1 & 100 & \/ & \/ & 4 & 100 & 4 \\
 & k\% & \/ & \/ & \/ & \/ & \/ & \/ & \/ & \/ & \/ & \/ & \/ & \/ & \/ & \/ & \/ \\
 & eachother & \/ & \/ & \/ & \/ & 7 & 100 & \/ & \/ & \/ & \/ & \/ & \/ & 7 & 100 & 7 \\
 & "unmatched" & 118 & 89.8 & 137 & 86.1 & 383 & 92.7 & 11 & 36.4 & 13 & 38.5 & 11 & 36.4 & 673 & 88 & 592 \\
 &  & \/ & \/ & \/ & \/ & \/ & \/ & \/ & \/ & \/ & \/ & \/ & \/ & \/ & \/ & \/ \\
between & some & \/ & \/ & \/ & \/ & \/ & \/ & 1 & 100 & \/ & \/ & \/ & \/ & 1 & 100 & 1 \\
 & all & \/ & \/ & \/ & \/ & \/ & \/ & \/ & \/ & \/ & \/ & \/ & \/ & \/ & \/ & \/ \\
 & k & \/ & \/ & 1 & 100 & \/ & \/ & 1 & 0 & 2 & 50 & \/ & \/ & 4 & 50 & 2 \\
 & between & \/ & \/ & 2 & 100 & \/ & \/ & 2 & 50 & \/ & \/ & \/ & \/ & 4 & 75 & 3 \\
 & most & \/ & \/ & \/ & \/ & \/ & \/ & \/ & \/ & \/ & \/ & \/ & \/ & \/ & \/ & \/ \\
 & few & \/ & \/ & \/ & \/ & \/ & \/ & \/ & \/ & \/ & \/ & \/ & \/ & \/ & \/ & \/ \\
 & \textgreater{}p/k:more/greaterthanp/k & \/ & \/ & \/ & \/ & \/ & \/ & \/ & \/ & \/ & \/ & \/ & \/ & \/ & \/ & \/ \\
 & \textless{}p/k:fewer/lessthanp/k & \/ & \/ & \/ & \/ & \/ & \/ & \/ & \/ & \/ & \/ & \/ & \/ & \/ & \/ & \/ \\
 & p/k & \/ & \/ & \/ & \/ & \/ & \/ & \/ & \/ & \/ & \/ & \/ & \/ & \/ & \/ & \/ \\
 & k\% & \/ & \/ & \/ & \/ & \/ & \/ & \/ & \/ & \/ & \/ & \/ & \/ & \/ & \/ & \/ \\
 & eachother & \/ & \/ & \/ & \/ & \/ & \/ & \/ & \/ & \/ & \/ & \/ & \/ & \/ & \/ & \/ \\
 & "unmatched" & 2 & 100 & 1 & 100 & \/ & \/ & 1 & 100 & \/ & \/ & \/ & \/ & 4 & 100 & 4 \\
 &  & \/ & \/ & \/ & \/ & \/ & \/ & \/ & \/ & \/ & \/ & \/ & \/ & \/ & \/ & \/ \\
most & some & \/ & \/ & 2 & 50 & \/ & \/ & \/ & \/ & \/ & \/ & \/ & \/ & 2 & 50 & 1 \\
 & all & \/ & \/ & 2 & 100 & \/ & \/ & \/ & \/ & \/ & \/ & \/ & \/ & 2 & 100 & 2 \\
 & \textgreater{}k & \/ & \/ & 1 & 0 & \/ & \/ & \/ & \/ & \/ & \/ & \/ & \/ & 1 & 0 & 0 \\
 & \textless{}k & \/ & \/ & \/ & \/ & \/ & \/ & \/ & \/ & \/ & \/ & \/ & \/ & \/ & \/ & \/ \\
 & k & 5 & 60 & 1 & 100 & \/ & \/ & \/ & \/ & 2 & 50 & 1 & 0 & 9 & 55.6 & 5 \\
 & between & \/ & \/ & \/ & \/ & \/ & \/ & \/ & \/ & \/ & \/ & \/ & \/ & \/ & \/ & \/ \\
 & most & 7 & 85.7 & 4 & 75 & \/ & \/ & \/ & \/ & \/ & \/ & \/ & \/ & 11 & 81.8 & 9 \\
 & few & 1 & 100 & \/ & \/ & \/ & \/ & \/ & \/ & \/ & \/ & \/ & \/ & 1 & 100 & 1 \\
 & p/k & \/ & \/ & \/ & \/ & \/ & \/ & \/ & \/ & \/ & \/ & \/ & \/ & \/ & \/ & \/ \\
 & k\% & \/ & \/ & \/ & \/ & \/ & \/ & \/ & \/ & \/ & \/ & \/ & \/ & \/ & \/ & \/ \\
 & eachother & \/ & \/ & \/ & \/ & \/ & \/ & \/ & \/ & \/ & \/ & \/ & \/ & \/ & \/ & \/ \\
 & "unmatched" & 23 & 73.9 & 30 & 83.3 & 1 & 0 & \/ & \/ & \/ & \/ & \/ & \/ & 54 & 77.8 & 42 \\
 &  & \/ & \/ & \/ & \/ & \/ & \/ & \/ & \/ & \/ & \/ & \/ & \/ & \/ & \/ & \/ \\
few & some & \/ & \/ & \/ & \/ & \/ & \/ & \/ & \/ & \/ & \/ & \/ & \/ & \/ & \/ & \/ \\
 & all & 1 & 100 & 1 & 0 & \/ & \/ & \/ & \/ & \/ & \/ & 0 &  & 2 & 50 & 1 \\
 & \textgreater{}k & \/ & \/ & 1 & 100 & \/ & \/ & \/ & \/ & \/ & \/ & 0 &  & 1 & 100 & 1 \\
 & \textless{}k & \/ & \/ & \/ & \/ & \/ & \/ & \/ & \/ & \/ & \/ & 0 &  & 0 &  & 0 \\
 & k & 4 & 75 & 3 & 33.3 & 4 & 100 & \/ & \/ & \/ & \/ & 3 & 33.3 & 14 & 64.3 & 9 \\
 & between & \/ & \/ & \/ & \/ & \/ & \/ & \/ & \/ & \/ & \/ & \/ & \/ & \/ & \/ & \/ \\
 & most & 0 &  & 1 & 0 & 0 &  & \/ & \/ & \/ & \/ & \/ & \/ & 1 & 0 & 0 \\
 & few & 0 &  & 2 & 100 & 0 &  & \/ & \/ & \/ & \/ & 1 & 0 & 3 & 66.7 & 2 \\
 & p/k & \/ & \/ & \/ & \/ & \/ & \/ & \/ & \/ & \/ & \/ & \/ & \/ & \/ & \/ & \/ \\
 & k\% & \/ & \/ & \/ & \/ & \/ & \/ & \/ & \/ & \/ & \/ & \/ & \/ & \/ & \/ & \/ \\
 & eachother & \/ & \/ & \/ & \/ & \/ & \/ & \/ & \/ & \/ & \/ & \/ & \/ & \/ & \/ & \/ \\
 & "unmatched" & 9 & 77.8 & 9 & 88.9 & 7 & 85.7 & \/ & \/ & \/ & \/ & 2 & 50 & 27 & 81.5 & 22 \\
 &  & \/ & \/ & \/ & \/ & \/ & \/ & \/ & \/ & \/ & \/ & \/ & \/ & \/ & \/ & \/ \\
p/k & some & \/ & \/ & \/ & \/ & \/ & \/ & \/ & \/ & \/ & \/ & \/ & \/ & \/ & \/ & \/ \\
 & all & \/ & \/ & \/ & \/ & \/ & \/ & \/ & \/ & \/ & \/ & \/ & \/ & \/ & \/ & \/ \\
 & \textgreater{}k & \/ & \/ & \/ & \/ & \/ & \/ & \/ & \/ & \/ & \/ & \/ & \/ & \/ & \/ & \/ \\
 & \textless{}k & \/ & \/ & \/ & \/ & \/ & \/ & \/ & \/ & \/ & \/ & \/ & \/ & \/ & \/ & \/ \\
 & k & \/ & \/ & 2 & 100 & 0 &  & 1 & 100 & 0 &  & 0 &  & 3 & 100 & 3 \\
 & between & \/ & \/ & \/ & \/ & \/ & \/ & \/ & \/ & \/ & \/ & \/ & \/ & \/ & \/ & \/ \\
 & most & \/ & \/ & \/ & \/ & \/ & \/ & \/ & \/ & \/ & \/ & \/ & \/ & \/ & \/ & \/ \\
 & few & \/ & \/ & \/ & \/ & \/ & \/ & \/ & \/ & \/ & \/ & \/ & \/ & \/ & \/ & \/ \\
 & p/k & \/ & \/ & 2 & 100 & \/ & \/ & 1 & 100 & \/ & \/ & \/ & \/ & 3 & 100 & 3 \\
 & k\% & \/ & \/ & \/ & \/ & \/ & \/ & \/ & \/ & \/ & \/ & \/ & \/ & \/ & \/ & \/ \\
 & eachother & \/ & \/ & \/ & \/ & \/ & \/ & \/ & \/ & \/ & \/ & \/ & \/ & \/ & \/ & \/ \\
 & "unmatched" & 1 & 100 & 3 & 33.3 & 1 & 100 & 0 &  & 1 & 100 & 0 &  & 6 & 66.7 & 4 \\
 &  & \/ & \/ & \/ & \/ & \/ & \/ & \/ & \/ & \/ & \/ & \/ & \/ & \/ & \/ & \/ \\
k\% & some & \/ & \/ & \/ & \/ & \/ & \/ & \/ & \/ & \/ & \/ & \/ & \/ & \/ & \/ & \/ \\
 & all & \/ & \/ & \/ & \/ & \/ & \/ & \/ & \/ & \/ & \/ & \/ & \/ & \/ & \/ & \/ \\
 & \textgreater{}k & \/ & \/ & \/ & \/ & \/ & \/ & \/ & \/ & \/ & \/ & \/ & \/ & \/ & \/ & \/ \\
 & \textless{}k & \/ & \/ & \/ & \/ & \/ & \/ & \/ & \/ & \/ & \/ & \/ & \/ & \/ & \/ & \/ \\
 & k & 6 & 83.3 & 6 & 100 & \/ & \/ & \/ & \/ & \/ & \/ & 1 & 100 & 13 & 92.3 & 12 \\
 & between & \/ & \/ & \/ & \/ & \/ & \/ & \/ & \/ & \/ & \/ & \/ & \/ & \/ & \/ & \/ \\
 & most & \/ & \/ & \/ & \/ & \/ & \/ & \/ & \/ & \/ & \/ & \/ & \/ & \/ & \/ & \/ \\
 & few & \/ & \/ & \/ & \/ & \/ & \/ & \/ & \/ & \/ & \/ & \/ & \/ & \/ & \/ & \/ \\
 & p/k & \/ & \/ & \/ & \/ & \/ & \/ & \/ & \/ & \/ & \/ & \/ & \/ & \/ & \/ & \/ \\
 & k\% & \/ & \/ & \/ & \/ & \/ & \/ & \/ & \/ & \/ & \/ & \/ & \/ & \/ & \/ & \/ \\
 & eachother & \/ & \/ & \/ & \/ & \/ & \/ & \/ & \/ & \/ & \/ & \/ & \/ & \/ & \/ & \/ \\
 & "unmatched" & 4 & 100 & 1 & 100 & \/ & \/ & \/ & \/ & \/ & \/ & \/ & \/ & 5 & 100 & 5 \\
 &  & \/ & \/ & \/ & \/ & \/ & \/ & \/ & \/ & \/ & \/ & \/ & \/ & \/ & \/ & \/ \\
eachother & some & \/ & \/ & \/ & \/ & \/ & \/ & \/ & \/ & \/ & \/ & 1 & 100 & 1 & 100 & 1 \\
 & all & \/ & \/ & \/ & \/ & 3 & 100 & \/ & \/ & \/ & \/ & \/ & \/ & 3 & 100 & 3 \\
 & \textgreater{}k & \/ & \/ & \/ & \/ & \/ & \/ & \/ & \/ & \/ & \/ & \/ & \/ & \/ & \/ & \/ \\
 & \textless{}k & \/ & \/ & \/ & \/ & \/ & \/ & \/ & \/ & \/ & \/ & \/ & \/ & \/ & \/ & \/ \\
 & k & 1 & 100 & \/ & \/ & 15 & 80 & \/ & \/ & \/ & \/ & 1 & 100 & 17 & 82.4 & 14 \\
 & between & \/ & \/ & \/ & \/ & \/ & \/ & \/ & \/ & \/ & \/ & \/ & \/ & \/ & \/ & \/ \\
 & most & \/ & \/ & \/ & \/ & \/ & \/ & \/ & \/ & \/ & \/ & \/ & \/ & \/ & \/ & \/ \\
 & few & \/ & \/ & \/ & \/ & \/ & \/ & \/ & \/ & \/ & \/ & \/ & \/ & \/ & \/ & \/ \\
 & p/k & \/ & \/ & \/ & \/ & \/ & \/ & \/ & \/ & \/ & \/ & \/ & \/ & \/ & \/ & \/ \\
 & k\% & \/ & \/ & \/ & \/ & \/ & \/ & \/ & \/ & \/ & \/ & \/ & \/ & \/ & \/ & \/ \\
 & eachother & \/ & \/ & \/ & \/ & 1 & 100 & \/ & \/ & \/ & \/ & \/ & \/ & 1 & 100 & 1 \\
 & "unmatched" & 3 & 66.7 & 3 & 100 & 19 & 89.5 & \/ & \/ & \/ & \/ & \/ & \/ & 26 & 84.6 & 22
\end{tabular}
}
\caption{Statistics of pairwise analysis in Monolingual NLI Benchmarks}
\label{tab:pairwise}
\end{table*}

%% file: tables/xqa_result_xmlr.tex
\begin{table*}[t]
\centering %\small 
% \scriptsize
% \footnotesize
\resizebox{\columnwidth}{!}{%
\begin{tabular}{lcccccccccccccc|cccccccccccccc}

\toprule

 \multirow{3}{*}{Quant.} & \multicolumn{14}{c}{mBERT} & \multicolumn{14}{c}{XLM-R} \\
  & \multicolumn{2}{c}{en} & \multicolumn{2}{c}{zh} & \multicolumn{2}{c}{es} & \multicolumn{2}{c}{ar} & \multicolumn{2}{c}{vi} & \multicolumn{2}{c}{de} & \multicolumn{2}{c}{\textit{weighted}} & \multicolumn{2}{c}{en} & \multicolumn{2}{c}{zh} & \multicolumn{2}{c}{es} & \multicolumn{2}{c}{ar} & \multicolumn{2}{c}{vi} & \multicolumn{2}{c}{de} & \multicolumn{2}{c}{\textit{weighted}}\\
 & EM & F1 & EM & F1 & EM & F1 & EM & F1 & EM & F1 & EM & F1 & EM & F1 & EM & F1 & EM & F1 & EM & F1 & EM & F1 & EM & F1 & EM & F1 & EM & F1 \\
  \midrule
\rowcolor{Gray}\multicolumn{29}{c}{XQuAD} \\
some &75 & 84.2 & 50 & 55.5 & 58.3 & 76.1 & 50 & \textbf{50} & 16.6 & 42.4 & \textbf{33.3} & \textbf{43.8} & 47.2 & 58.7 & \textbf{66.7} & \textbf{76.1} & \textbf{41.6} & \textbf{51.3} & \textbf{50} & \textbf{71.5} & 66.7 & 73.6 & 66.7 & \textbf{76.9} & 66.7 & 80.6 & 59.7 & 71.7\\
all & \textbf{28.5} & \textbf{62.2} & \textbf{14.2} & \textbf{35.2} & \textbf{28.5} & 82 & 42.8 & \textbf{52.3} & \textbf{14.2} & \textbf{29.4} & \textbf{28.5} & \textbf{56} & \textbf{26.1} & \textbf{52.9} & \textbf{57.1} & 91.8 & \textbf{14.2} & \textbf{21.4} & \textbf{57.1} & \textbf{78.6} & \textbf{42.8} & \textbf{54.9} & 85.7 & 85.7 & \textbf{57.1} & 79.3 & \textbf{52.3} & \textbf{68.6} \\
\(>k\) & &  &  &  &  &  &  &  &  &  &  &  &  &  &  &  &  &  &  &  &  &  &  &  &  &  &     \\
\(<k\) & &  &  &  &  &  &  &  &  &  &  &  &  &  &  &  &  &  &  &  &  &  &  &  &  &  &     \\
\(k\) &78.1 & 90.1 & 68.7 & 80.4 & 56.2 & \textbf{72.1} & \textbf{40.6} & 64.3 & \textbf{12.5} & 35.7 & 56.2 & 77.1 & 52.1 & 70 & 75 & 87.4 & 53.1 & 58.8 & \textbf{46.8} & \textbf{77.4} & 65.6 & 86.3 & 62.5 & 85.4 & 62.5 & 86.9 & 60.9 & 80.4\\
between & 100 & 100 & \textbf{33.3} & \textbf{72.2} & 66.6 & 93.3 & 100 & 100 & \textbf{0} & \textbf{19} & \textbf{0} & \textbf{56.5} & 50 & 73.5 & 100 & 100 & 66.7 & 66.7 & \textbf{33.3} & \textbf{60} & 100 & 100 & 100 & 100 & \textbf{33.3} & \textbf{55.5} & 72.2 & 80.4\\
\(p/k\) &  &  &  &  &  &  &  &  &  &  &  &  &  &  &  &  &  &  &  &  &  &  &  &  &  &  &     \\
\(k \%\) &  &  &  &  &  &  &  &  &  &  &  &  &  &  &  &  &  &  &  &  &  &  &  &  &  &  &     \\
most & \textbf{40} & \textbf{53.3} & \textbf{40} & \textbf{40} & \textbf{0} & \textbf{10} & \textbf{0} & \textbf{26.6} & \textbf{0} & \textbf{0} & \textbf{20} & \textbf{49.3} & \textbf{16.7} & \textbf{29.9} & \textbf{40} & \textbf{48} & \textbf{20} & \textbf{33.3} & \textbf{40} & \textbf{50} & \textbf{0} & \textbf{26.6} & \textbf{0} & \textbf{0} & \textbf{20} & \textbf{49.3} & \textbf{20} & \textbf{34.5}\\
few &  &  &  &  &  &  &  &  &  &  &  &  &  &  &  &  &  &  &  &  &  &  &  &  &  &  &     \\
each other &  &  &  &  &  &  &  &  &  &  &  &  &  &  &  &  &  &  &  &  &  &  &  &  &  &  &     \\
\midrule
all GQs & \textbf{70} & \textbf{83.2} & 55 & 66.7 & \textbf{50} & \textbf{70.3} & 41.6 & 58.2 & \textbf{11.6} & 32.5 & \textbf{43.3} & \textbf{65} & \textbf{45.3} & 62.7 & \textbf{70} & \textbf{83.6} & \textbf{43.3} & \textbf{50.2} & \textbf{48.3} & \textbf{73.6} & 60 & 76 & 68.3 & 83.6 & \textbf{58.3} & 80.3 & \textbf{58} & 74.6\\
comp. & 71.8 & 83.7 & 48 & 59.1 & 56 & 74.5 & 40.8 & 57.9 & 13.9 & 32.4 & 50.7 & 67.2 & 46.9 & 62.5 & 74.5 & 86 & 43 & 52.8 & 61 & 80 & 53.3 & 71.7 & 58.1 & 78 & 61.1 & 77.1 & 58.5 & 74.3\\
\bottomrule
\rowcolor{Gray}\multicolumn{29}{c}{MLQA} \\
some & \textbf{59} & 80 & \textbf{28.2} & \textbf{52.1} & \textbf{34.1} & \textbf{59.2} & 36.3 & 54.9 & \textbf{5.4} & \textbf{24} & \textbf{33.3} & 58.4 & \textbf{32.7} & \textbf{54.8} & \textbf{69.6} & 86.1 & \textbf{33.3} & \textbf{60.6} & \textbf{41.4} & \textbf{70} & 43.1 & 62.9 & \textbf{43.2} & 78 & \textbf{45.4} & \textbf{61.1} & \textbf{46} & \textbf{69.8}\\
 all & 67.7 & \textbf{79.8} & \textbf{14.2} & \textbf{46.4} & \textbf{38.4} & \textbf{62.8} & 33.3 & 57.9 & \textbf{10.5} & 30.1 &\textbf{31.2} & \textbf{51.6} & \textbf{32.6} & \textbf{54.8} & 77.4 & 90.6 & \textbf{35.7} & 70 & \textbf{42.3} & \textbf{66.4} & \textbf{38} & \textbf{60} & 57.8 & 79.8 & \textbf{37.5} & \textbf{51} & \textbf{48.1} & \textbf{69.6}\\
\(>k\) &  &  &  &  &  &  &  &  &  &  &  &  &  &  &  &  &  &  &  &  &  &  &  &  &  &  &  & \\
\(<k\) & 0 & 0 &  &  &  &  &  &  & 0 & 13.3 &  &  & 0 & 6.7 & 0 & 40 &  &  &  &  &  &  & 0 & 20 &  &  & 0 & 30\\
\(k\) & \textbf{74.9} & 79.4 & 47 & 63.4 & \textbf{41.5} & \textbf{65.9} & \textbf{27.6} & 50.3 & \textbf{6.3} & \textbf{23.7} & \textbf{38.2} & \textbf{53} & 39.3 & \textbf{56} & \textbf{69.2} & \textbf{82.1} & 45.2 & 66.2 & \textbf{48.7} & 73.3 & 43 & 64.9 & \textbf{48.5} & \textbf{71.9} & \textbf{46.3} & \textbf{62.1} & \textbf{50.2} & 70.1\\
between & \textbf{50} & 88.5 & 50 & 83.3 & \textbf{0} & \textbf{26.6} & \textbf{0} & 68.7 & \textbf{0} & \textbf{26.6} &  &  & \textbf{20} & 58.7 & \textbf{50} & 88.5 & 50 & \textbf{50} & \textbf{50} & \textbf{65.3} & \textbf{0} & \textbf{54.6} & \textbf{0} & 77.4 &  &  & \textbf{30} & \textbf{67.2}\\
\(p/k\) & 100 & 100 & \textbf{0} & \textbf{0} & \textbf{0} & \textbf{0} &  &  &  &  &  &  & \textbf{33.3} & \textbf{33.3} & 100 & 100 & 100 & 100 & 100 & 100 &  &  &  &  &  &  & 100 & 100\\
\(k \%\) & 100 & 100 & \textbf{0} & \textbf{26.6} &  &  & \textbf{0} & \textbf{23.7} &  &  &  &  & \textbf{33.3} & \textbf{50.1} & 100 & 100 & 0 & \textbf{26.6} &  &  & \textbf{0} & 71.4 &  &  &  &  & \textbf{33.3} & \textbf{66}\\
most & \textbf{55.5} & \textbf{7} & 47.3 & 62.1 & \textbf{45.4} & \textbf{61.7} & 30 & \textbf{46.8} & \textbf{5.8} & \textbf{15.7} & \textbf{33.3} & \textbf{40.7} & \textbf{36.2} & \textbf{50.3} & \textbf{59.2} & \textbf{76} & 47.3 & 69.5 & \textbf{45.4} & \textbf{59.5} & \textbf{40} & 63.2 & \textbf{47} & 75.7 & \textbf{22.2} & \textbf{31.7} & \textbf{43.5} & \textbf{62.6}\\
few & &  &  &  &  &  &  &  &  &  &  &  &  & & &  &  &  &  &  &  &  &  &  &  &  &  &\\
each other & &  &  &  &  &  &  &  &  &  &  &  &  & & &  &  &  &  &  &  &  &  &  &  &  &  & \\
\midrule
all GQs & \textbf{63.5} & \textbf{79.2} & 41.8 & 60.3 & \textbf{39.6} & \textbf{63.7} & \textbf{29.3} & 51.3 & \textbf{6.4} & \textbf{23.6} & \textbf{36.1} & \textbf{53.2} & \textbf{36.1} & \textbf{55.2} & \textbf{69} & \textbf{83} & 43 & 65.6 & \textbf{46.9} & \textbf{71.5} & \textbf{41.9} & 64.1 & \textbf{47.6} & 73.2 & \textbf{44.4} & \textbf{59.8} & \textbf{48.8} & \textbf{69.5}\\
comp. & 67.2 & 80.6 & 37.5 & 57.9 & 47.3 & 66 & 30 & 48.4 & 11.2 & 28 & 40.8 & 56 & 39 & 56.2 & 70.4 & 83.3 & 38.7 & 62.5 & 54.1 & 72.2 & 42.5 & 62.9 & 50.5 & 72.3 & 52.2 & 67.3 & 51.4 & 70.1\\
\bottomrule

\end{tabular}
}
\caption{Results of mBERT and XLM-R performance on XQA tasks decomposed by quantifier categories.}
\label{tab:xqa_both_results}
\end{table*}

%% file: tables/gqnli_examples.tex
\begin{table*}[t]
\footnotesize

\centering %\small 
% \scriptsize
\small 
\begin{tabular}{lp{7cm}p{3.8cm}c}

\toprule

\textbf{Quantifier} & \textbf{Premise} & \textbf{Hypothesis} & \textbf{Label} \\
\midrule
some & ``There are six dogs. Three brown dogs, a black dog and a white dog run along the green grass.'' & ``Some dogs sit.'' & Neutral \\
all & ``In 2021, there are 490 million people in Africa living in extreme poverty, or 36\% of the total population.'' & ``Not all people in Africa live in extreme poverty.'' & Entailment \\
\(>k\) & ``Two young men in blue stand over a stove and look at the camera while another young man in red stands behind them.'' & ``At least two men wear red.'' & Contradiction \\
\(<k\) & ``More than five guys chased two girls in the classroom.'' & ``No less than four guys chased two girls in the classroom.'' & Entailment \\
\(k\)  &  ``There are twelve singers on a stage, less than half from Argentina and one from Cape Verde.'' & ``Two singers come from Argentina.'' & Neutral \\
between & ``Only half out of six cleaners are sweeping up animal faeces from the street during a parade.'' & ``Between four and five cleaners are sweeping up animal faeces.'' & Contradiction  \\
\(p/k\) & ``More than 50\% but less than 65\% of Americans worry about global warming.'' & ``Two thirds of Americans worry about global warming.'' & Contradiction \\
\(k \%\) & ``More than five guys chased two girls in the classroom.'' & ``100\% of the guys chased two girls in the classroom.'' & Neutral\\
most & ``Two young men in blue stand over a stove and look at the camera while another young man in red stands behind them.'' & ``Most men wear blue.'' & Entailment \\
 few & ``More than 50\% but less than 65\% of Americans worry about global warming.'' & ``A few people from America do not worry about global warming.'' & Entailment \\
 each other & ``There are 100 villagers and 100 townsmen. Most villagers and most townsmen hate each other.'' & ``All villagers and all townsmen hate each other.'' & Neutral\\

\bottomrule
\end{tabular}
\caption{GQNLI examples.}
\label{tab:gqnli_example}
\end{table*}

%% file: tables/x_negation.tex
\begin{table*}[t]
    \small
    \resizebox{\textwidth}{!}{%
    \begin{tabular}{llccccccccccc|c}
    \toprule
    
    \multicolumn{2}{c}{\textbf{Quantifier}} & some & all & \(>k\) & \(<k\) & k & between & \(p/k\) & \(k \%\) & most & few & each other & \textit{Overall} \\ 
     \multicolumn{2}{c}{\textbf{\# Occurrence with negation cues}} & 9 & 6 & 6 & 9 & 18 & 3 & 6 & 6 & 6 & 9 & 3 & 81 \\ \midrule
    \textbf{Model} & \textbf{Training Data} & \multicolumn{12}{c}{\textbf{\% Performance}} \\ \midrule
\multirow{1}{*}{BERT} & S,M,F,ANLI & 0 & 66.7 & 100 & 33.3 & 50 & 0 & 50 & 0 & 50 & 22.2 & 33.3 & 39.2 \\ \midrule
 \multirow{1}{*}{ELECTRA} & S,M,F,ANLI & 33.3 & 50.0 & 100.0 & 33.3 & 50.0 & 0.0 & 50.0 & 0.0 & 66.7 & 0.0 & 0.0 & 43.1 \\ \midrule
 \multirow{1}{*}{SBERT} & S,M,F,ANLI & 55.6 & 50.0 & 66.7 & 11.1 & 27.8 & 0.0 & 50.0 & 0.0 & 100.0 & 66.7 & 0.0 & 54.9 \\ \midrule
 \multirow{2}{*}{RoBERTa} & MNLI & 33.3 & 16.7 & 0 & 33.3 & 27.8 & 66.7 & 33.3 & 33.3 & 50 & 33.3 & 33.3 & 31.4\\ 
     & S,M,F,ANLI & 66.7 & 83.3 & 100.0 & 33.3 & 66.7 & 100.0 & 50.0 & 50.0 & 50.0 & 33.3 & 66.7 & 58.8 \\ \midrule
\multirow{1}{*}{ALBERT} & S,M,F,ANLI & 88.9 & 50.0 & 66.7 & 33.3 & 55.6 & 100.0 & 0.0 & 50.0 & 50.0 & 11.1 & 0.0 & 49.0 \\ \midrule
 \multirow{2}{*}{BART} & MNLI & 33.3 & 0.0 & 50.0 & 66.7 & 66.7 & 100.0 & 0.0 & 100.0 & 0.0 & 33.3 & 0.0 & 35.3 \\ 
     & S,M,F,ANLI & 66.7 & 50.0 & 100.0 & 33.3 & 50.0 & 0.0 & 50.0 & 0.0 & 50.0 & 66.7 & 100.0 & 52.9\\ \midrule
\multirow{3}{*}{DeBERTa-v3} & MNLI & 33.3 & 0.0 & 50.0 & 33.3 & 50.0 & 100.0 & 66.7 & 50.0 & 0.0 & 33.3 & 0.0 & 37.3 \\ 
     & M,F,ANLI & 55.6 & 66.7 & 100.0 & 33.3 & 66.7 & 100.0 & 50.0 & 50.0 & 100.0 & 55.6 & 33.3 & 66.7 \\ 
     & M,F,Ling,DocNLI & 33.3 & 100.0 & 100.0 & 0.0 & 33.3 & 0.0 & 83.3 & 0.0 & 50.0 & 66.7 & 100.0 & 51.0 \\ \bottomrule
     \end{tabular}
     }
\caption{Models' performance on instances with negation cues in GQNLI.}
\label{tab:gqnli_neg}
\end{table*}

%% file: tables/subsumption_result.tex
\begin{table*}[t]
    \small
    \resizebox{\textwidth}{!}{%
    \begin{tabular}{llccc|c}
    \toprule
    
    \multicolumn{2}{c}{\textbf{Type}} & Subsumption (neutral) & Subsumption (non-neutral) & Subsumption (total) & Non-subsumption \\ 
     \multicolumn{2}{c}{\textbf{\# Occurrence}} & 90 & 99 & 189 & 111 \\ \midrule
    \textbf{Model} & \textbf{Training Data} & \multicolumn{4}{c}{\textbf{\% Performance}} \\ \midrule
 \multirow{1}{*}{BERT} & S,M,F,ANLI &  22.2 & 24.2 & 23.3 & 41.4 \\ \midrule
 \multirow{1}{*}{ELECTRA} & S,M,F,ANLI & 3.3 & 52.5 & 29.1 & 53.2 \\ \midrule
 \multirow{1}{*}{SBERT} & S,M,F,ANLI & 68.9 & 35.4 & 51.3 & 18.9\\ \midrule
 \multirow{2}{*}{RoBERTa} & MNLI & 27.8 & 18.2 & 22.8 & 37.8  \\ 
     & S,M,F,ANLI & 21.1 & 33.3 & 27.5 & 59.5\\ \midrule
\multirow{1}{*}{ALBERT} & S,M,F,ANLI & 33.3 & 38.4 & 36.0 & 49.5 \\ \midrule
 \multirow{2}{*}{BART} & MNLI & 36.7 & 46.5 & 41.8 & 40.5 \\ 
     & S,M,F,ANLI & 44.4 & 23.2 & 33.3 & 58.6  \\ \midrule
\multirow{3}{*}{DeBERTa-v3} & MNLI & 45.6 & 26.3 & 35.4 & 33.3\\ 
     & M,F,ANLI & 52.2 & 37.4 & 44.4 & 54.1  \\ 
     & M,F,Ling,DocNLI & 86.7 & 17.2 & 50.3 & 36.0\\ \bottomrule
     \end{tabular}
     }
\caption{Models' performance on instances requiring subsumption reasoning.}
\label{tab:gqnli_sub}
\end{table*}

%% file: tables/huggingface.tex
\begin{table*}[t]
    \small
    \resizebox{\textwidth}{!}{%
    \begin{tabular}{lll}
    \toprule
    \textbf{Model} & \textbf{Training Data} & \multicolumn{1}{c}{\textbf{Model's link}} \\ \midrule
 \multirow{1}{*}{ELECTRA} & S,M,F,ANLI & \url{https://huggingface.co/ynie/electra-large-discriminator-snli_mnli_fever_anli_R1_R2_R3-nli}\\ \midrule
 \multirow{1}{*}{SBERT} & S,M,F,ANLI & \url{https://huggingface.co/usc-isi/sbert-roberta-large-anli-mnli-snli} \\ \midrule
 \multirow{2}{*}{BART} & MNLI & \url{https://huggingface.co/facebook/bart-large-mnli} \\ 
     & S,M,F,ANLI & \url{https://huggingface.co/ynie/bart-large-snli_mnli_fever_anli_R1_R2_R3-nli}  \\ \midrule
\multirow{1}{*}{ALBERT} & S,M,F,ANLI & \url{https://huggingface.co/ynie/albert-xxlarge-v2-snli_mnli_fever_anli_R1_R2_R3-nli}\\ \midrule
\multirow{3}{*}{DeBERTa-v3} & MNLI & \url{https://huggingface.co/MoritzLaurer/DeBERTa-v3-base-mnli} \\ 
     & M,F,ANLI & \url{https://huggingface.co/MoritzLaurer/DeBERTa-v3-base-mnli-fever-anli} \\ 
     & M,F,Ling,DocNLI & \url{https://huggingface.co/MoritzLaurer/DeBERTa-v3-base-mnli-fever-docnli-ling-2c}\\ \bottomrule
     \end{tabular}
     }
\caption{Links to the models we use to test on GQNLI.}
\label{tab:huggingface}
\end{table*}